\pdfoutput=1
\documentclass{article}
\usepackage[preprint]{neurips_2024}
\usepackage{amsmath,amsfonts,bm}

\def\eqref#1{equation~\ref{#1}}
\def\1{\bm{1}}

\DeclareMathAlphabet{\mathsfit}{\encodingdefault}{\sfdefault}{m}{sl}
\SetMathAlphabet{\mathsfit}{bold}{\encodingdefault}{\sfdefault}{bx}{n}

\usepackage[utf8]{inputenc}
\usepackage[T1]{fontenc}
\usepackage{hyperref}
\usepackage{url}
\usepackage{graphicx}
\usepackage{booktabs}
\usepackage{xcolor}
\usepackage{enumitem}
\usepackage{microtype}
\usepackage{caption}
\usepackage{float}
\usepackage[capitalize,noabbrev]{cleveref}
\graphicspath{{figs/}}
\hypersetup{colorlinks=true,linkcolor=black,citecolor=blue!50!black,urlcolor=blue!50!black,
  pdftitle={How Language Models Differ in Redistributing Attention-Head Activity Under Serial Demand},
  pdfauthor={Johnny Jingze Li, Abdulla Kuleib, Kalyan Basu, Gabriel A. Silva}}

\title{How Language Models Differ in\\ Redistributing Attention-Head Activity\\ Under Serial Demand}

\author{%
  Johnny Jingze Li$^{1,2}$\thanks{Corresponding author: \texttt{jil164@ucsd.edu}}\quad
  Abdulla Kuleib$^{2}$\quad
  Kalyan Basu$^{2,3}$\quad
  Gabriel A. Silva$^{2}$\\[0.4em]
  \textmd{$^{1}$Carnegie Mellon University\quad
  $^{2}$University of California, San Diego\quad
  $^{3}$Qualtrics LLC}
}

\newcommand{\topq}{\mathrm{S}^{(0.1)}}

\begin{document}
\maketitle

\begin{abstract}
The way a model distributes activity over each layer's attention heads offers a coarse view of how it
routes information through depth; how this changes with the task is part of what a mechanistic account
must explain. Holding prompt length fixed, we vary how many serial steps a task demands and measure, in
every layer of 17 open-weight models, whether activity concentrates on a few heads or spreads across many
as demand rises. Both occur: in most models, layers just before mid-depth concentrate activity and later layers spread it. Models differ in where and how strongly this happens. The Qwen2.5 base models from 0.5B to 7B, for
example, spread less than the average model in every task and concentrate activity in parts of their
second half, where Llama models from 1B to 8B and OLMo-2 spread; the contrast largely holds between Llama-3.1-70B and
Qwen2.5-72B, which have the same number of layers and heads. These differences are reproducible,
and post-trained models keep much of their base model's pattern. An ablation study suggests that,
within a task, models whose activity is more concentrated on their top heads also depend more on those
heads for the answer. Concentration and spreading across layers thus offer a new way to compare models,
by how they route information through depth. Code is available at \url{https://github.com/johnnyjli/serial-demand-heads}.
\end{abstract}

\section{Introduction}

Large language models (LLMs) perform a fixed number of layer computations per token. When a
question requires composing several steps (following a pointer $k$ times, resolving a chain of
variable bindings, applying an operation repeatedly) and no intermediate tokens are produced, the
network must do whatever it can within one forward pass. Fixed-depth transformers are provably
limited in how many serial steps they can compose \citep{sanford2024,merrill2024}, and models often
fail to compose facts they know individually \citep{press2023,biran2024,yang2024}. How the network
reallocates its internal activity as serial demand rises is much less understood, and to our
knowledge no study has asked whether models differ in this respect.

Most interpretability results come from a single model and task
\citep[e.g.,][]{wang2023ioi,hanna2023}. Whether they transfer to other models is the question of
\emph{universality} \citep{olah2020,chughtai2023,gurnee2024}. We ask it for a simple, model-agnostic
quantity that can be measured identically across many models: how a layer's activity is distributed
over its attention heads, as given by the norms of the heads' outputs. A layer \emph{concentrates}
when its activity moves onto a few strong heads as serial demand grows, and \emph{spreads} when its
activity is distributed more evenly. This distribution is a coarse view of information flow: heads move
information between positions by writing into the residual stream \citep{elhage2021}, and circuit
analyses explain a behavior through the few heads that carry it \citep{wang2023ioi,conmy2023,ferrando2024}.
How it changes with the number of steps bears on how sparse these circuits are, and on whether a circuit
found on a simple instance describes a harder one. We ask two questions. \textbf{Do models differ} in which layers
spread or concentrate as the number of required steps $k$ grows? And \textbf{do those differences hold
across tasks}, i.e., are they properties of the model rather than of a particular task?

\paragraph{Contributions.}
\begin{itemize}[leftmargin=1.4em,itemsep=1pt,topsep=2pt]
\item \textbf{A controlled cross-model protocol} (\cref{sec:setup}): four serial task families with a step count $k$, constant prompt length and balanced answers; a verified per-head measurement at the
answer position in 17 models; and a decomposition of each model's depth profile into a part shared by
all models, a task part, a model part and a model-by-task part.
\item \textbf{Reproducible model differences, partly shared across tasks} (\cref{sec:signatures}):
depth profiles replicate across item halves, and a model-specific component recurs across tasks well
above chance, though weakly between tasks that do and do not write $k$. Most model-specific structure,
however, is task-specific.
\item \textbf{How models differ} (\cref{sec:differ}): in whether and where they spread after
mid-depth, whether a concentrating band precedes the spreading, and how consistent their response is
across tasks.
\item \textbf{Lineage and correlates} (\cref{sec:lineage,sec:track}): post-trained models keep much of
their base model's profile; the main axis of difference is associated with accuracy and with heads per
layer but persists after adjusting for answer entropy and among models with equal head counts.
\item \textbf{Common tendencies} (\cref{sec:shared}): most models spread activity after mid-depth,
and post-trained models tend to spread more than their base models, a tendency that is not specific
to serial demand.

\end{itemize}

\section{Related work}
\label{sec:related}

\paragraph{Attention heads and their activity.} Heads differ widely in importance; many can be
pruned with little loss \citep{voita2019,michel2019}, and some implement identifiable operations
such as induction \citep{olsson2022}. \citet{kobayashi2020} showed that attention weights alone
misrepresent a head's influence and analyzed the norms of the weighted, transformed value vectors
instead; we track the norm of each head's output in the same spirit. Heads that attend
mostly to the first token \citep[attention sinks;][]{xiao2024} and outlier heads with very large
outputs, analogous to the massive activations of \citet{sun2024}, can distort norm-based analyses; we
control for both.

\paragraph{Serial and multi-hop computation.} Chain of thought relaxes the limits of fixed depth \citep{sanford2024,merrill2024} by externalizing steps
\citep{wei2022}; without it, models often fail to combine two known facts \citep{press2023}, and the second hop is resolved late or not at all \citep{biran2024,yang2024}. \citet{csordas2025} found that harder problems do not recruit more layers. We study the within-layer counterpart: how heads share activity as $k$ grows.

\paragraph{Universality and model comparison.} Comparisons across networks have found shared
representations \citep{kornblith2019,huh2024} and features or neurons that recur across training runs
\citep{olah2020,gurnee2024}, alongside run-to-run variation in how a task is implemented
\citep{chughtai2023}. Layer-wise studies describe processing stages common to many LLMs
\citep{lad2024}. Fine-tuning often reuses the base model's mechanisms
\citep{jain2024,prakash2024}, consistent with our finding that post-trained models keep much of their
base model's profile.

\section{Setup}
\label{sec:setup}

\subsection{Serial tasks}
\label{sec:tasks}

The four families compose one repeated operation a controllable number of times, each with a
single-digit answer and an exact ground truth, and they are of two types: two write the number of steps in the prompt and two do not. Each family has a depth parameter $k$, the number of steps the prompt specifies
(\cref{tab:tasks}; prompts in \cref{app:tasks}). In pointer chasing, binding and relational lookup, $k$ is also the fewest lookups that determine the answer, although transformers can compose lookups in parallel \citep{sanford2024}. \emph{Pointer chasing} uses a random 10-cycle with
$k\le5$, since on an $n$-cycle $n-k$ steps back reach the state $k$ steps ahead. \emph{Modular
arithmetic} uses modulus 7; its answer $(s+kd)\bmod7$, for start $s$ and increment $d$, has a closed form, so there $k$ need not add
sequential computation (\cref{app:tasks}).
\emph{Variable binding} and \emph{relational lookup} interleave, in shuffled order, three chains of
seven statements each (21 in all), each rooted in a different digit. The prompt body is identical
across $k$ and only the queried entity changes, so these two families never write $k$ and have the same answer set, length and average position of the queried statement at every depth
(\cref{app:tasks}). Answers are balanced within each depth. Each item has three
worked examples at depths 1--3, none of which is the item's own query. Each family has two wordings that share instances. A
\emph{numeral control} keeps the written step count at 7 while $k$ sets the length ($k+1$) of the
cycle through the start. The fewest lookups needed are 1, 1, 1, 2, 1 and 0 for $k=1,\dots,6$, so $k$
does not increase the number of steps. An independent solver re-derived the answers of all 2{,}436 prompts
and 7{,}308 worked examples from the prompt text alone, without a mismatch.
Of the 60 items generated per depth and wording, we use the 42 test items (for modular arithmetic,
24 distinct problems per depth, repeated with different worked examples).

\begin{table}[t]
\centering\small
\caption{Task families. Prompt length is constant across $k$ within a family. Chance is one over the
number of admissible answers.}
\label{tab:tasks}
\setlength{\tabcolsep}{4.5pt}
\begin{tabular}{@{}lp{7.6cm}ccc@{}}
\toprule
Family & Query (wording w0) & $k$ & $k$ written & Chance \\
\midrule
Pointer chasing & \small``Rules: 0 goes to 5. \ldots\ Start at 4. Take $k$ steps. You end at'' & 1--5 & yes & 0.10 \\
Modular arithmetic & \small``Rule: add 2 and keep the remainder after dividing by 7. Start at 3. Apply the rule $k$ times. You end at'' & 1--6 & yes & 0.14 \\
Variable binding & \small``Assignments: w = z. v = n. \ldots\ d = 6. \ldots\ Therefore z ='' & 1--6 & no & 0.33 \\
Relational lookup & \small``Chart: The boss of M is G. \ldots\ The top boss above Z is'' & 1--6 & no & 0.33 \\
\midrule
Numeral control & \small pointer chasing on 7 states; written step count fixed at 7 & 1--6 & fixed & 0.14 \\
\bottomrule
\end{tabular}
\end{table}

\subsection{Models}

We study 17 decoder-only models (\cref{tab:models}): Llama-3.2-1B and 3B \citep{llama32} and
Llama-3.1-8B \citep{llama3} with the 3B and 8B Instruct models; Qwen2.5 from 0.5B to 14B
\citep{qwen25} with the 7B Instruct model; OLMo-2-1124-7B with its SFT, DPO and Instruct checkpoints
\citep{olmo2}; and Mistral-7B-v0.1 \citep{mistral} and Mistral-Small-24B-Base-2501
\citep{mistralsmall3}. They form 11 \emph{lineages}, each a base model from one pretraining run with any
post-trained versions of it, in four \emph{families} (developers). All models receive the same raw
prompts.

\begin{table}[t]
\centering\small
\caption{Models: $L$ layers, $H$ query heads per layer. Some figures and tables shorten OLMo-2-1124-7B to
OLMo-2-7B and Mistral-Small-24B to Mistral-24B.}
\label{tab:models}
\begin{tabular}{@{}llcc@{\hspace{1.6em}}llcc@{}}
\toprule
Lineage & Stages & $L$ & $H$ & Lineage & Stages & $L$ & $H$ \\
\midrule
Llama-3.2-1B & base & 16 & 32 & Qwen2.5-3B & base & 36 & 16 \\
Llama-3.2-3B & base, Instruct & 28 & 24 & Qwen2.5-7B & base, Instruct & 28 & 28 \\
Llama-3.1-8B & base, Instruct & 32 & 32 & Qwen2.5-14B & base & 48 & 40 \\
Qwen2.5-0.5B & base & 24 & 14 & OLMo-2-1124-7B & base, SFT, DPO, Instruct & 32 & 32 \\
Qwen2.5-1.5B & base & 28 & 12 & Mistral-7B-v0.1 & base & 32 & 32 \\
 & & & & Mistral-Small-24B & base & 40 & 32 \\
\bottomrule
\end{tabular}
\end{table}

\subsection{Measuring head activity}
\label{sec:measure}

For each item we run one forward pass and read every layer at the \emph{answer position}, the token
whose next-token prediction is the answer digit (for tokenizers that split `` 6'' into a space and a
digit, we append the space token). For head $h$ in layer $\ell$, with attention-weighted value vector
$v_{\ell,h}$ and output-projection slice $W_O^{(h)}$, its activity is $s_{\ell,h}=\|v_{\ell,h}W_O^{(h)}\|_2$,
the norm of what it writes toward the residual stream. On every run we check that the per-head writes
sum to the output projection's output (relative error below $5\times10^{-2}$ in bf16). Models run in
bf16 with PyTorch's standard (eager) attention, except Qwen2.5-7B-Instruct, which uses fused scaled dot-product attention (SDPA) because its eager-attention
outputs are numerically degraded. We captured its base model with SDPA in only one task and wording and analyze it under eager attention (\cref{app:robust}). In OLMo-2, we exclude items in which one head outputs more
than 50 times its layer's median head norm in either wording: up to 64\% of the arithmetic
items for the base model, at most 18 of 252 binding or relational items and no pointer items. Apart from
the sign of OLMo-2's post-training change (\cref{sec:shared}), no result depends on the exclusion
(\cref{tab:robust} in \cref{app:robust}).

\paragraph{Concentration statistic.} With shares $p_{\ell,h}=s_{\ell,h}/\sum_{h'}s_{\ell,h'}$ sorted
in decreasing order $p_{\ell,(1)}\ge p_{\ell,(2)}\ge\dots$, the \emph{top-10\% head share} is
\begin{equation}
\topq_\ell=\sum_{j=1}^{\lceil 0.1H\rceil}p_{\ell,(j)},
\end{equation}
the fraction of the layer's activity carried by its strongest tenth of heads (two heads when
$H\le16$). A rising $\topq_\ell$ means the layer \emph{concentrates} activity onto its strongest heads
as $k$ grows; a falling one means it \emph{spreads} activity more evenly. We report the top-25\% and
top-50\% shares, the effective number of heads ($\exp$ of the entropy of $p_{\ell,\cdot}$) and the
share of the single strongest head as alternatives.

\subsection{Statistics}
\label{sec:stats}

\paragraph{Per-layer response.} We average each instance's statistic over its two wordings. For each model, task family and layer we regress $\topq_\ell$ on $k$ across items and
standardize the slope by its permutation distribution, $z_\ell=\beta_\ell/\mathrm{sd}(\beta^\pi_\ell)$,
where $\pi$ shuffles $k$ across items (pointer, arithmetic) or within prompt bodies (binding,
relational), one shuffle for all layers (2{,}000 shuffles). Negative $z_\ell$ means the layer spreads as $k$
grows, positive that it concentrates. A layer is significant if its two-sided permutation $p<0.05$. The \emph{back half} comprises the layers at or beyond 50\% of relative depth $\ell/(L-1)$.

\paragraph{Profiles and signatures.} To compare models with different depths we average $z_\ell$
within relative-depth deciles, giving a profile $Z_{m,t,d}$ for model $m$, task $t$ and decile $d$
($17\times4\times10$). We decompose
\begin{equation}
Z_{m,t,d}=\mu_d+a_{m,d}+b_{t,d}+e_{m,t,d},
\end{equation}
where $\mu$ is the profile averaged over all models and tasks, $a_m$ is model $m$'s \emph{signature}
(its deviation from $\mu$, averaged over its four tasks), $b_t$ is the task effect and $e$ the
model-by-task interaction. We assess the signatures in four ways. (i) \emph{Label permutation}:
shuffling model labels independently within each task keeps every task's profiles but destroys model
identity across tasks; the size of $a$ under 5{,}000 shuffles is its chance level. (ii)
\emph{Split-half}: we correlate signatures estimated from two tasks with those from the other two,
for each of the three splits. (iii) \emph{Item-split reliability}: we recompute all profiles on
disjoint halves of the instances (pointer, arithmetic) or prompt bodies (binding, relational) in four
random splits. (iv) \emph{Similarity}: the correlation of two models' signatures.

\section{Results}
\label{sec:results}

\begin{figure}[t]
\centering
\includegraphics[width=\textwidth]{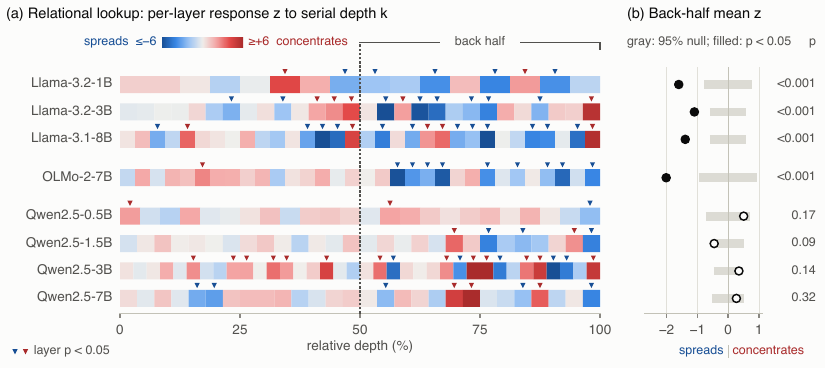}
\caption{\textbf{In relational lookup, the back half of the Llama models and OLMo-2 spreads activity
as $k$ grows; that of the Qwen2.5 models up to 7B shows no significant spreading.} (a) Standardized slope $z$ of each
layer's \mbox{top-10\%} head share on $k$ (blue: spreads; red: concentrates; clipped at $\pm6$); triangles: $p<0.05$, uncorrected. (b) Mean $z$ over back-half layers; gray: 95\% range of its permutation null; filled: $p<0.05$. Permutations shuffle $k$ within prompt bodies. Base models
only. Eager attention may understate spreading in Qwen2.5-7B (\cref{app:robust}); Qwen2.5-14B (not shown) spreads like the Llama models ($-2.06$, $p<0.001$).}
\label{fig:contrast}
\end{figure}

\subsection{Models differ reproducibly, and partly consistently across tasks}
\label{sec:signatures}

\Cref{fig:contrast} shows the clearest difference we find: in relational lookup, the back half of each
Llama base model and of OLMo-2 spreads activity as $k$ grows, whereas that of each Qwen2.5 base model up
to 7B shows no significant spreading (\cref{sec:differ}). \Cref{fig:profiles} in \cref{app:figs} shows every model's
depth profile on each task. The profiles are reliable:
recomputed on disjoint halves of the items, a model's profile on a task correlates $r=0.84$ with
itself (0.74--0.92 by task), and the model signatures replicate at $r=0.93$. For arithmetic this reliability is somewhat inflated, as its problems repeat with different worked examples. On fresh items, newly generated for a rerun of eight models (\cref{app:replication}), a model's signature correlates $r=0.80$ on average with its signature
here ($p=2.5\times10^{-5}$), and seven of the eight match their own signature better
than any other model's.

The average profile $\mu$ accounts for 18\% of the variation in $Z$. Of the remainder, model signatures account for 34\% (21\% by chance, $p<0.001$), task effects for 15\% (4\% by chance), and the
model-by-task interaction for 51\%, which is not noise: it replicates across item halves ($r=0.84$). Corrected for the number of models and tasks, the variance components of model, task and
interaction are 0.17, 0.15 and 0.67, so most of what distinguishes a model on a given task is specific
to that task. What recurs is still well above chance:
signatures estimated from two tasks predict those from the other two (split-half $r=0.36$; chance
0.00, $p<0.001$), and a model on two different tasks resembles itself more ($r=0.30$) than different
models on the same task ($r=0.22$). On fresh items, model identity explains 29\% of the non-shared
variation (19\% by chance, $p=0.003$), with split-half $r=0.31$.

How far a model's deviation carries over depends on the tasks. Between tasks that do and do not write
$k$ it agrees weakly ($r=0.03$--0.20; 0.14 on fresh items; predicting binding and relational lookup from the other two gives $r=0.17$, against 0.45 and 0.46 for the other splits). It agrees more within
each type, between pointer chasing and arithmetic ($r=0.32$) and between binding and relational lookup
($r=0.54$), but the latter falls to 0.25 on fresh items, whose prompts for these tasks differ, and we do not
rely on it (\cref{app:replication}). Consistency also differs by model: OLMo-2 (all four checkpoints;
$r=0.57$--0.62) and Mistral-7B (0.60) have the most consistent profiles across tasks (no other model
exceeds 0.34), whereas the full profiles of the three smallest Qwen2.5 models barely correlate across
tasks ($-0.01$ to 0.06), although their weak spreading relative to $\mu$ recurs in every task.

\begin{figure}[t]
\centering
\includegraphics[width=\textwidth]{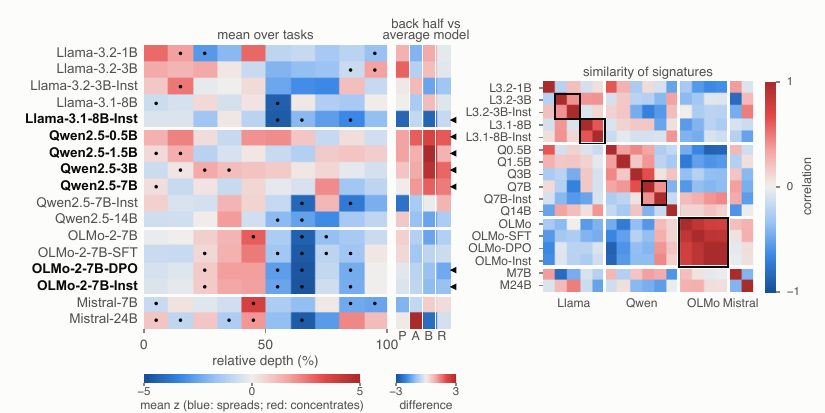}
\caption{\textbf{Some models spread more, or less, than the others in every task.} \emph{Left:} each
model's profile averaged over the four tasks (blue: spreading; red: concentrating; dots: deciles where all
four tasks have the same sign), and, per task (P, A, B, R: pointer chasing, arithmetic, binding,
relational lookup), its back-half response minus that of the average model (blue: spreads more; red:
less). Bold names and triangles: the same sign in all four tasks. \emph{Right:} correlation between model signatures $a_m$. Black squares: lineages.}
\label{fig:signatures}
\end{figure}

\subsection{How the models differ}
\label{sec:differ}

\paragraph{Some models spread more, or less, than the others in every task.} In the back half, seven
of the 17 models deviate from the average model in the same direction in all four tasks, against 1.8
expected when model labels are shuffled within each task ($p<0.001$; strip in \cref{fig:signatures}).
The four Qwen2.5 base models up to 7B spread less than the average model in every task and spread at
all in only one or two of the four; averaged over tasks, those up to 3B do not spread in the back half
(mean $z$ from $+0.1$ to $+0.4$), and Qwen2.5-7B spreads only slightly ($-0.2$, under eager attention, which may understate its response; \cref{app:robust}). Llama-3.1-8B-Instruct and the OLMo-2
DPO and Instruct checkpoints spread more than the average model in every task, and at 50--70\% of depth
all four OLMo-2 checkpoints do (seven models at that depth, against 2.1 expected, $p=0.004$). Ten models,
among them Llama-3.2-1B, Llama-3.1-8B and all four OLMo-2 checkpoints, spread in the back half in
every task, and in relational lookup, arithmetic and binding these four Qwen2.5 models separate completely from the three Llama base models (exact permutation, $p=1/35$ in each task; \cref{fig:contrast}), though not in pointer chasing. The
contrast largely holds between Llama-3.1-70B and Qwen2.5-72B, which have the same number of layers and
heads: in three of the four tasks, the former spreads in the back half, whereas the latter changes little
on net and concentrates at 80--90\% of depth in all four (\cref{app:large}). The same-direction deviations concern the strongest heads: they recur under the share of the single strongest head (nine models, $p<0.001$) but not under the top-25\%
and top-50\% shares or the effective number of heads, under all of which only Qwen2.5-0.5B and 1.5B spread less than average in every task (\cref{app:robust}). On fresh items, with one wording, no rerun model keeps one sign in all four tasks, although Qwen2.5-7B again spreads least of
them at 50--80\% of depth (\cref{app:replication}).

\paragraph{Differences that vary more between tasks.} The task-averaged profiles (\cref{fig:signatures}, left; \cref{tab:layers,fig:layermaps}) show four further
differences. \emph{Where models spread}: Llama-3.1-8B sharply at 50--60\% of depth (layers 16--17),
OLMo-2 and Mistral-Small-24B at 60--70\%, Qwen2.5-7B-Instruct at 60--70\% and 80--90\%, Llama-3.2-3B and
Qwen2.5-14B across 50--80\%, and Mistral-7B at 80--100\%, whereas Qwen2.5-1.5B, 3B and 7B have layers
that concentrate consistently at 54--81\%; on fresh items, the decile of strongest spreading recurs in
seven of the eight rerun models (2.5 expected, $p=0.004$). \emph{Whether a concentrating band precedes
the spreading}: the OLMo-2 base model and both Mistral models concentrate at 40--50\% of depth in all
four tasks (on fresh items, again in OLMo-2 but not in Mistral-Small-24B), Llama-3.2-1B and the
post-trained OLMo-2 checkpoints in pointer chasing and arithmetic only, and Llama-3.2-3B, Llama-3.1-8B
and Qwen2.5-7B-Instruct in a single layer (at 44\%, 48\% and 44\%). \emph{How localized the response
is}: OLMo-2's strongest spreading sits in a block of adjacent layers (18--21 of 32), whereas
Llama-3.1-8B-Instruct's consistently spreading layers are scattered from 6\% to 97\% of depth.
\emph{What the final layer does}: averaged over tasks, it spreads in all six Qwen2.5 models and has positive $z$ in 10 of the other 11 (all but Llama-3.2-1B; only $+0.02$ in Mistral-7B; label permutation $p<0.001$, or 0.013 with one value per lineage), a pattern we found while exploring that recurs on fresh items in
all eight rerun models ($p=1/56$). It does not continue the back-half profile ($r=0.07$ with the 80--90\% decile), and the other families' concentration depends on the statistic
(5 to 8 of 11 under other statistics), whereas Qwen2.5's spreading does not.

\begin{figure}[t]
\centering
\includegraphics[width=\textwidth]{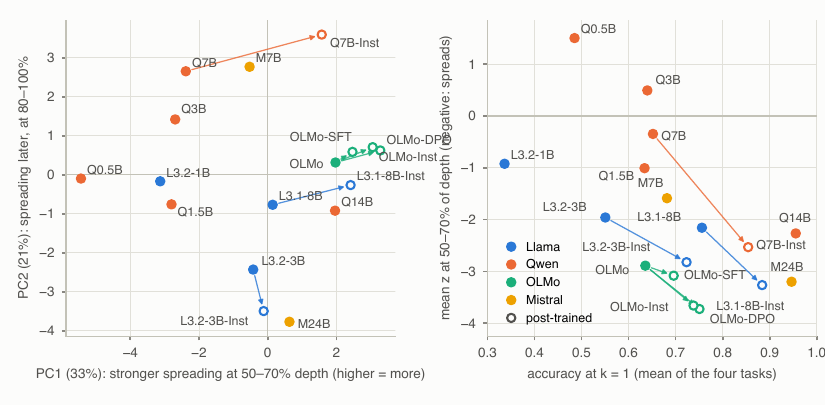}
\caption{\textbf{Axes of difference.} \emph{Left:} model signatures on their first two principal
components. \emph{Right:} mean $z$ at 50--70\% of depth, averaged over the four tasks, against accuracy
at $k=1$. Color: family; open markers: post-trained models, with arrows from their base model. L: Llama;
Q: Qwen2.5; M: Mistral.}
\label{fig:axes}
\end{figure}

The first two principal components of the signatures (\cref{fig:axes}, left) capture the first of
these differences: PC1 (33\% of signature variance) is the strength of spreading at 50--70\% of depth
(correlation $-0.93$ with the 50--70\% signature), and PC2 (21\%) is whether spreading occurs at
70--80\% or at 80--100\%. The concentrating band lies mostly on the third and fourth components.
Consistent layers, significant in the same direction in at least three tasks and in the opposite
direction in none, cluster by depth: 40 of the 56 spreading ones lie at 50--90\% (15 expected by chance, $p<0.001$;
\cref{tab:layers}), and 9 of the 20
concentrating ones at 35--50\% (1.6 expected, $p<0.001$); of the 14 layers that spread consistently on fresh items, 11 also do so here (about 3 expected within the back half).

\subsection{Post-trained models keep much of their base model's profile}
\label{sec:lineage}

A base model and its post-trained versions have signatures that correlate $r=0.64$ on average: 0.40 for
Llama-3.2-3B, 0.62 for Llama-3.1-8B, 0.40 for Qwen2.5-7B and 0.76--0.87 for the three OLMo-2
checkpoints (permutation $p<0.001$); all nine within-lineage pairs, six of them among OLMo-2
checkpoints, average 0.73, consistent with evidence that fine-tuning reuses base-model mechanisms
\citep{jain2024,prakash2024}. Models from different pretraining runs are uncorrelated (mean $r=-0.11$,
against $-0.06$ expected from centering; 0.06 for other sizes from the same developer and
$-0.15$ across developers), and among the 11 base models, those from the same developer are only
marginally more similar than chance (0.09 against $-0.05$, $p=0.06$), echoing run-to-run differences in circuits and
neurons \citep{chughtai2023,gurnee2024}. On fresh items, the three rerun
lineage pairs again have similar signatures (mean $r=0.64$) and models from different pretraining runs
do not ($-0.23$; $-0.14$ expected from centering), although OLMo-2's pair, the most similar here, is the least similar there (0.50). The OLMo-2
checkpoints keep the same consistently spreading layers (18, 19 and 21), and Llama-3.1-8B and its
Instruct version share 16, 17 and 24 (\cref{tab:layers}). Retention is partial: the
signatures closest to those of Llama-3.2-3B-Instruct and Qwen2.5-7B-Instruct belong to
Mistral-Small-24B and OLMo-2-Instruct, not to their base models. Within Qwen2.5, the four smaller base
models have similar signatures (mean $r=0.39$), whereas Qwen2.5-14B does not ($r=-0.43$ with
Qwen2.5-7B; $-0.49$ on fresh items). Its signature does not replicate across tasks
(split-half $r=-0.06$) but does on fresh items (0.41), where it correlates 0.82 with its signature
here. As head count grows with size within this family, we cannot tell whether size, head count or
pretraining data decides this. Post-training also tends to shift profiles in a common direction
(\cref{sec:shared}).

\subsection{What the differences are associated with}
\label{sec:track}

The main axis of difference, PC1, is associated with both accuracy and heads per layer (\cref{fig:axes},
right). Across the 17 models, PC1 correlates with mean accuracy (Spearman $\rho=0.78$) and heads per layer (0.64). We summarize the axis by the \emph{50--70\% response}, a model's mean $z$ over the 50--60\%
and 60--70\% deciles averaged over tasks, which differs from its 50--70\% signature by a constant common
to all models. The response recurs on fresh items ($r=0.82$ across the eight rerun models, $p=0.005$;
\cref{app:replication}) and correlates with mean accuracy ($\rho=-0.73$, $p=0.001$), accuracy
at $k=1$ ($-0.65$), the growth with $k$ of answer entropy over the admissible digits ($-0.64$) and heads per layer ($-0.57$,
$p=0.02$). Among the 11 base models, PC1 still tracks mean accuracy ($\rho=0.81$, $p=0.004$), and the
response tracks accuracy ($-0.70$, $p=0.02$) and head count ($-0.68$, $p=0.03$) about equally. Head count is confounded here with family and size: the three weakest responses belong to Qwen2.5-0.5B, 3B
and 7B ($+1.5$, $+0.5$ and $-0.3$), with 14--28 heads per layer, and every model with $H\ge32$ spreads ($z\le-0.9$). The Qwen2.5 pattern does not come from the coarse top-10\% group: under the effective number of heads, the four Qwen2.5 base
models up to 7B also spread less at 50--70\% of depth ($+1.3$ to $-1.0$) than any other
family's model ($-1.9$ to $-4.2$).

Neither factor accounts for the differences alone. Signatures differ among models with equal head
count: among the nine with $H=32$, model identity explains 30\% of the non-shared variation (18\%
by chance, $p<0.001$; 27\% against 19\% among the five base models, $p=0.006$), and the 50--70\% response
still follows mean accuracy ($\rho=-0.83$, $p=0.008$; $-0.89$, $p=0.03$ without the post-trained OLMo-2
checkpoints). Nor does equal accuracy imply equal signatures: Qwen2.5-3B and OLMo-2-7B both answer 64\%
of $k=1$ items correctly, but their responses are $+0.5$ and $-2.9$. The link with accuracy survives adjusting
each item for answer entropy: the response then correlates $-0.72$ with mean accuracy
($p=0.0015$), and the adjusted PC1 still tracks it ($\rho=0.73$, $p=0.001$).

\paragraph{Dependence on the most active heads.} Our ablation study suggests that, within a task,
models whose activity is more concentrated depend more on their most active heads. At the answer
position, we mean-ablated \citep{wang2023ioi} the tenth of heads with the largest mean activity in every layer at 40--80\%
of depth and compared the loss of answer margin with that from as many random mid-ranked heads
(\cref{app:ablation}). The excess drop rises with the level of a model's top-10\% share: Spearman $\rho$ across models is 0.36--0.75 per task (mean 0.59, permutation $p<0.001$;
\cref{fig:ablation}). The association rests mainly on the Qwen2.5 models up to 7B, which have low shares and whose most active heads are no more necessary than random ones; within families it is weak,
and Qwen2.5-14B depends strongly on its top heads despite a low share. A model's share barely changes
between tasks, so the ablation cannot test the relation across tasks within a model.

\subsection{Robustness of the signatures}
\label{sec:robust}

The signatures survive changes to the statistic (the \mbox{top-50\%} share agrees least, $r=0.42$), attention-sink
heads \citep{xiao2024}, the attention kernel, the OLMo-2 exclusion, depth binning or near-duplicate
post-trained models: model identity explains 29--36\% of non-shared variation in every variant,
against 18--23\% by chance (\cref{tab:robust}). Signatures depend more on wording than on item sampling (\cref{app:robust}). Accuracy decreases with $k$, as expected for composition in one forward pass \citep{sanford2024,merrill2024}, but at a given $k$ head activity is nearly the same for
correct and incorrect answers, and adjusting for correctness changes the signatures little,
so we use all items (\cref{app:accuracy}).

\subsection{What the models share}
\label{sec:shared}

The differences sit on top of a common profile ($\mu$; \cref{fig:profiles}, dashed) and a few secondary
tendencies.

\paragraph{Spreading after mid-depth.} Weighting each of the 11 lineages once, layers at 40--50\% of
depth concentrate (mean $z=+1.19$, 95\% CI across lineages $[0.42, 2.02]$, mostly from arithmetic) and
layers at 50--80\% spread (CIs exclude zero); the 80--100\% deciles are negative, but their CIs include
zero. The back half spreads more than the front half in 10 of 11 lineages (16 of 17 models) and spreads
outright in 8 (14 of 17 models). On fresh items (\cref{app:replication}), layers at 50--80\% again spread
(lineage-weighted mean $z=-1.02$, 95\% CI $[-1.70,-0.34]$), and the back half of all eight rerun models
spreads outright and more than the front half. The drop at mid-depth is itself common: the 40--50\%
decile lies above the 50--60\% decile in 16 of 17 models (10 of 11 lineages), and in 12 of 17 the sign
changes from concentrating to spreading. We found this while exploring; no other pair of adjacent
deciles shows a drop as often (next: 13 of 17 models, 9 of 11 lineages). This change lies near the phase
transition between the two halves of the network reported by \citet{csordas2025} and the
prediction-ensembling stage of \citet{lad2024}. The back half spreads more in
the serial tasks than in the numeral control, where $k$ does not add steps, in 13 of 17 models; the base
models do not spread in the control on average, although answer entropy also grows less with $k$ there.
Deeper items are also harder, so serial demand and difficulty cannot be fully separated; the control,
which also becomes harder with $k$ (\cref{tab:acc}), and the entropy adjustment separate them in part.
 The
shared spreading is small: per step, the top-10\% share falls by 0.10 percentage points (95\% CI
0.06--0.13; 0.09 on fresh items), against a baseline back-half share of 25--51\% across models; a few heads thus carry much of each layer's output \citep[cf.][]{voita2019,michel2019}. Adjusting for the model's growing answer entropy removes about half of this fall (\cref{app:shared}).

\paragraph{Post-trained models tend to spread more.} In three of the four lineages, the post-trained
model's back half spreads more as $k$ grows than its base model's: the mean back-half $z$, averaged over
tasks, changes by $-0.76$ for Llama-3.2-3B, $-1.49$ for Llama-3.1-8B and $-1.57$ for Qwen2.5-7B. The three OLMo-2 checkpoints show no net change ($+0.03$ on average,
$-0.25$ to $+0.45$ depending on which items the exclusion removes; \cref{fig:posttrain}). On fresh items,
all three rerun post-trained models spread more than their bases ($-0.73$, $-0.46$ and $-0.96$ for
Llama-3.1-8B, Qwen2.5-7B and OLMo-2-Instruct), so the direction recurs while its size varies between runs. Every post-trained model also lies further along PC1 than its base
(\cref{fig:axes}), and on identical items its top-10\% share at 40--80\% of depth falls faster with $k$
in all four lineages, though barely for OLMo-2. We treat this as a tendency, for three reasons
(\cref{app:shared}). First, the 95\% $t$-interval for the lineage-weighted change, $[-2.14, 0.25]$,
includes zero, and the Qwen2.5-7B pair mixes attention kernels. Second, the shift goes together with
confidence: every post-trained model is more accurate at $k=1$, and adjusting each item for its answer
entropy removes the shift in both Llama pairs. Third, the shift is not specific to serial demand: in the numeral control, the post-trained models spread more
than their bases in all four lineages, by at least as much as in the serial tasks, even after the entropy adjustment.

\paragraph{Larger models tend to spread more at 50--70\% of depth.} In all 14 size-ordered pairs of
base models from the same family, the larger model lies further along PC1 (exact within-family
permutation $p<0.001$), but ten pairs are Qwen2.5 models, three of the other four also differ in model
generation, and size and accuracy are rank-correlated at $\rho=0.94$ among base models, so we cannot
attribute this to size. Over the whole back half, the larger model spreads more only within Qwen2.5.

\section{Discussion}
\label{sec:discussion}

\paragraph{Differences are reproducible but only partly general.} Models differ reliably in how
they redistribute head activity, across item halves and on fresh items, but only part of each model's deviation carries over to a task of a different type. A
layer-level finding from one model and task, such as ``layers 18--21 respond to serial demand,'' is
therefore most likely to transfer to the same model on a similar task and to its post-trained
descendants, and least likely to other models or kinds of task.

\paragraph{Post-training preserves profiles; pretraining runs differ.} Post-trained models keep much of their
base model's profile, by an amount that varies between lineages and runs, whereas separate pretraining
runs are uncorrelated, even from the same developer. The post-trained models' stronger spreading appears in the numeral control too, so we read it as a response to harder items rather than to serial steps. With 11 lineages we cannot decide what the pretraining runs differ
in: accuracy and heads per layer, both associated with the main axis, are confounded with family and
size in our sample.

\paragraph{What spreading and concentration mean for information flow and reasoning.} A layer's head activity at the
answer position shows how many heads carry what it writes into the residual stream, which carries information to the prediction \citep{elhage2021,geva2023}. A layer that
concentrates as $k$ grows routes more of its output through a few heads, like the sparse circuits found
for single behaviors \citep{wang2023ioi,hanna2023}; a layer that spreads shares it more evenly, although in the back half this happens because the strongest heads lose relative weight as total output falls, not because more heads are recruited (\cref{app:shared}). For reasoning in one forward pass, spreading after mid-depth has two readings we cannot yet separate: layers just past mid-depth may commit less
to one answer as the chain lengthens, since the main axis tracks growing answer entropy and many
errors stop part of the way along the chain (\cref{app:accuracy}; cf.\ \citealp{biran2024}), or models
may compose steps partly in parallel across heads rather than one per layer \citep{sanford2024}. Either way, the heads carrying a multi-step computation depend on the model, and a circuit found at small $k$
may be less concentrated at larger $k$, so circuit analyses of reasoning should be checked across values of $k$ and models. Ablation gives the
level of concentration a functional counterpart: in models where more of a layer's activity runs
through a few heads, the answer depends more on those heads, mostly as a difference between families (\cref{sec:track}). It does not show what spreading layers compute as $k$ grows, which needs interventions at earlier positions or activation patching.

\section{Conclusion}

Language models differ reproducibly in how they redistribute attention-head activity as a task
demands more serial steps: in whether and where they spread it after mid-depth, whether they
concentrate it before, and how consistently they do so. Part of each model's pattern recurs across
tasks, though weakly across task types, and post-trained models keep much of their base model's pattern.
Beneath these differences, most models spread activity after mid-depth, more when $k$ adds steps than
when it does not, and post-trained models tend to spread more than their base models in both cases.

\bibliography{refs}

@inproceedings{biran2024,
  title={Hopping Too Late: Exploring the Limitations of Large Language Models on Multi-Hop Queries},
  author={Biran, Eden and Gottesman, Daniela and Yang, Sohee and Geva, Mor and Globerson, Amir},
  booktitle={Proceedings of the 2024 Conference on Empirical Methods in Natural Language Processing},
  year={2024}
}

@article{csordas2025,
  title={Do Language Models Use Their Depth Efficiently?},
  author={Csord{\'a}s, R{\'o}bert and Manning, Christopher D. and Potts, Christopher},
  journal={arXiv preprint arXiv:2505.13898},
  year={2025}
}

@article{llama3,
  title={The {L}lama 3 Herd of Models},
  author={Grattafiori, Aaron and others},
  journal={arXiv preprint arXiv:2407.21783},
  year={2024}
}

@inproceedings{jain2024,
  title={Mechanistically Analyzing the Effects of Fine-Tuning on Procedurally Defined Tasks},
  author={Jain, Samyak and Kirk, Robert and Lubana, Ekdeep Singh and Dick, Robert P. and Tanaka, Hidenori and Grefenstette, Edward and Rockt{\"a}schel, Tim and Krueger, David Scott},
  booktitle={International Conference on Learning Representations},
  year={2024}
}

@article{mistral,
  title={Mistral {7B}},
  author={Jiang, Albert Q. and others},
  journal={arXiv preprint arXiv:2310.06825},
  year={2023}
}

@inproceedings{kobayashi2020,
  title={Attention is Not Only a Weight: Analyzing Transformers with Vector Norms},
  author={Kobayashi, Goro and Kuribayashi, Tatsuki and Yokoi, Sho and Inui, Kentaro},
  booktitle={Proceedings of the 2020 Conference on Empirical Methods in Natural Language Processing},
  year={2020}
}

@article{lad2024,
  title={The Remarkable Robustness of {LLM}s: Stages of Inference?},
  author={Lad, Vedang and Gurnee, Wes and Tegmark, Max},
  journal={arXiv preprint arXiv:2406.19384},
  year={2024}
}

@inproceedings{merrill2024,
  title={The Expressive Power of Transformers with Chain of Thought},
  author={Merrill, William and Sabharwal, Ashish},
  booktitle={International Conference on Learning Representations},
  year={2024}
}

@inproceedings{michel2019,
  title={Are Sixteen Heads Really Better than One?},
  author={Michel, Paul and Levy, Omer and Neubig, Graham},
  booktitle={Advances in Neural Information Processing Systems},
  year={2019}
}

@article{olmo2,
  title={2 {OLMo} 2 {Furious}},
  author={{Team OLMo}},
  journal={arXiv preprint arXiv:2501.00656},
  year={2024}
}

@article{olsson2022,
  title={In-context Learning and Induction Heads},
  author={Olsson, Catherine and others},
  journal={Transformer Circuits Thread},
  year={2022}
}

@inproceedings{prakash2024,
  title={Fine-Tuning Enhances Existing Mechanisms: A Case Study on Entity Tracking},
  author={Prakash, Nikhil and Shaham, Tamar Rott and Haklay, Tal and Belinkov, Yonatan and Bau, David},
  booktitle={International Conference on Learning Representations},
  year={2024}
}

@inproceedings{press2023,
  title={Measuring and Narrowing the Compositionality Gap in Language Models},
  author={Press, Ofir and Zhang, Muru and Min, Sewon and Schmidt, Ludwig and Smith, Noah A. and Lewis, Mike},
  booktitle={Findings of the Association for Computational Linguistics: EMNLP 2023},
  year={2023}
}

@article{qwen25,
  title={Qwen2.5 Technical Report},
  author={{Qwen Team}},
  journal={arXiv preprint arXiv:2412.15115},
  year={2024}
}

@inproceedings{sanford2024,
  title={Transformers, Parallel Computation, and Logarithmic Depth},
  author={Sanford, Clayton and Hsu, Daniel and Telgarsky, Matus},
  booktitle={International Conference on Machine Learning},
  year={2024}
}

@inproceedings{sun2024,
  title={Massive Activations in Large Language Models},
  author={Sun, Mingjie and Chen, Xinlei and Kolter, J. Zico and Liu, Zhuang},
  booktitle={Conference on Language Modeling},
  year={2024}
}

@inproceedings{voita2019,
  title={Analyzing Multi-Head Self-Attention: Specialized Heads Do the Heavy Lifting, the Rest Can Be Pruned},
  author={Voita, Elena and Talbot, David and Moiseev, Fedor and Sennrich, Rico and Titov, Ivan},
  booktitle={Proceedings of the 57th Annual Meeting of the Association for Computational Linguistics},
  year={2019}
}

@inproceedings{wang2023ioi,
  title={Interpretability in the Wild: a Circuit for Indirect Object Identification in {GPT-2} Small},
  author={Wang, Kevin and Variengien, Alexandre and Conmy, Arthur and Shlegeris, Buck and Steinhardt, Jacob},
  booktitle={International Conference on Learning Representations},
  year={2023}
}

@inproceedings{wei2022,
  title={Chain-of-Thought Prompting Elicits Reasoning in Large Language Models},
  author={Wei, Jason and Wang, Xuezhi and Schuurmans, Dale and Bosma, Maarten and Ichter, Brian and Xia, Fei and Chi, Ed and Le, Quoc V. and Zhou, Denny},
  booktitle={Advances in Neural Information Processing Systems},
  year={2022}
}

@inproceedings{xiao2024,
  title={Efficient Streaming Language Models with Attention Sinks},
  author={Xiao, Guangxuan and Tian, Yuandong and Chen, Beidi and Han, Song and Lewis, Mike},
  booktitle={International Conference on Learning Representations},
  year={2024}
}

@inproceedings{yang2024,
  title={Do Large Language Models Latently Perform Multi-Hop Reasoning?},
  author={Yang, Sohee and Gribovskaya, Elena and Kassner, Nora and Geva, Mor and Riedel, Sebastian},
  booktitle={Proceedings of the 62nd Annual Meeting of the Association for Computational Linguistics},
  year={2024}
}

@article{gurnee2024,
  title={Universal Neurons in {GPT2} Language Models},
  author={Gurnee, Wes and Horsley, Theo and Guo, Zifan Carl and Kheirkhah, Tara Rezaei and Sun, Qinyi and Hathaway, Will and Nanda, Neel and Bertsimas, Dimitris},
  journal={Transactions on Machine Learning Research},
  year={2024}
}

@inproceedings{chughtai2023,
  title={A Toy Model of Universality: Reverse Engineering how Networks Learn Group Operations},
  author={Chughtai, Bilal and Chan, Lawrence and Nanda, Neel},
  booktitle={International Conference on Machine Learning},
  year={2023}
}

@inproceedings{huh2024,
  title={The {P}latonic Representation Hypothesis},
  author={Huh, Minyoung and Cheung, Brian and Wang, Tongzhou and Isola, Phillip},
  booktitle={International Conference on Machine Learning},
  year={2024}
}

@inproceedings{kornblith2019,
  title={Similarity of Neural Network Representations Revisited},
  author={Kornblith, Simon and Norouzi, Mohammad and Lee, Honglak and Hinton, Geoffrey},
  booktitle={International Conference on Machine Learning},
  year={2019}
}

@article{olah2020,
  title={Zoom In: An Introduction to Circuits},
  author={Olah, Chris and Cammarata, Nick and Schubert, Ludwig and Goh, Gabriel and Petrov, Michael and Carter, Shan},
  journal={Distill},
  year={2020}
}

@inproceedings{hanna2023,
  title={How does {GPT-2} compute greater-than?: Interpreting mathematical abilities in a pre-trained language model},
  author={Hanna, Michael and Liu, Ollie and Variengien, Alexandre},
  booktitle={Advances in Neural Information Processing Systems},
  year={2023}
}

@misc{llama32,
  title={Llama 3.2 Model Card},
  author={{Meta AI}},
  year={2024},
  howpublished={\url{https://github.com/meta-llama/llama-models/blob/main/models/llama3_2/MODEL_CARD.md}}
}

@misc{mistralsmall3,
  title={{Mistral Small 3}},
  author={{Mistral AI}},
  year={2025},
  howpublished={\url{https://mistral.ai/news/mistral-small-3}}
}

@article{elhage2021,
  title={A Mathematical Framework for Transformer Circuits},
  author={Elhage, Nelson and Nanda, Neel and Olsson, Catherine and Henighan, Tom and Joseph, Nicholas and Mann, Ben and Askell, Amanda and Bai, Yuntao and Chen, Anna and Conerly, Tom and DasSarma, Nova and Drain, Dawn and Ganguli, Deep and Hatfield-Dodds, Zac and Hernandez, Danny and Jones, Andy and Kernion, Jackson and Lovitt, Liane and Ndousse, Kamal and Amodei, Dario and Brown, Tom and Clark, Jack and Kaplan, Jared and McCandlish, Sam and Olah, Chris},
  journal={Transformer Circuits Thread},
  year={2021},
  note={https://transformer-circuits.pub/2021/framework/index.html}
}

@inproceedings{conmy2023,
  title={Towards Automated Circuit Discovery for Mechanistic Interpretability},
  author={Conmy, Arthur and Mavor-Parker, Augustine N. and Lynch, Aengus and Heimersheim, Stefan and Garriga-Alonso, Adri{\`a}},
  booktitle={Advances in Neural Information Processing Systems},
  year={2023}
}

@inproceedings{ferrando2024,
  title={Information Flow Routes: Automatically Interpreting Language Models at Scale},
  author={Ferrando, Javier and Voita, Elena},
  booktitle={Proceedings of the 2024 Conference on Empirical Methods in Natural Language Processing},
  pages={17432--17445},
  year={2024}
}

@inproceedings{geva2023,
  title={Dissecting Recall of Factual Associations in Auto-Regressive Language Models},
  author={Geva, Mor and Bastings, Jasmijn and Filippova, Katja and Globerson, Amir},
  booktitle={Proceedings of the 2023 Conference on Empirical Methods in Natural Language Processing},
  year={2023}
}
\bibliographystyle{plainnat}

\clearpage
\appendix
\raggedbottom

\section{Task details}
\label{app:tasks}

\paragraph{Generators.} \emph{Pointer chasing} draws a random single 10-cycle over the digits, a start
state and $k\in\{1,\dots,5\}$. The answer is the state $k$ steps ahead and is balanced over the nine
digits other than $k$. \emph{Modular arithmetic} draws a start $s$ and an increment $d$, and the answer
is $(s+kd)\bmod 7$. \emph{Variable binding} and \emph{relational lookup} interleave three chains of
seven statements with digit roots, in shuffled order. Items with the same body share every statement
at all $k$ and differ only in the queried entity, whose distance to its root is $k$. The \emph{numeral
control} has 7 states and a written step count of 7, and its table is one cycle of length $k+1$ through
the start, with all other states as fixed points. Each item has three worked examples at depths 1--3,
none of which is its own query. In binding and relational lookup they are drawn per prompt body and
shared by its six items ($k=1,\dots,6$).

\paragraph{What $k$ measures.} The depth $k$ is the number of steps the prompt specifies. In pointer chasing,
every table is a single 10-cycle and $k\le5$, so the fewest lookups that determine the answer equal $k$;
in binding and relational lookup, each of the $k$ statements between the queried entity and its root is
needed, and no other statement links them. A transformer need not perform these lookups one per layer:
$k$ lookups can be composed in parallel with depth logarithmic in $k$ \citep{sanford2024}, so $k$
measures how much composition an answer needs rather than how many layers it uses. Modular arithmetic
has the closed form $(s+kd)\bmod7$, which a model could compute with one multiplication instead of $k$
additions. The models' near-chance accuracy at $k\ge2$ and the many intermediate values among their errors (\cref{app:accuracy}) suggest that they rarely do so, but the closed form may contribute to
arithmetic's weaker back-half response (\cref{app:shared}).

\paragraph{Positions in the chain tasks.} Across the items of both wordings, the statement that defines
the queried entity sits at the same average position at every depth (Spearman correlation with $k$ of
$-0.03$ in binding and $-0.01$ in relational lookup), and so does the root statement (0.00). What grows
with $k$ is the span of the statements a query needs, from none at $k=1$ to about 16 of the 21 at $k=6$
($\rho=0.76$ and 0.78), because a deeper query needs more of them. In a within-body regression of each
item's back-half top-10\% share on $k$, adding the position of the queried statement leaves the
response unchanged ($|t|>1.96$ and negative in 49 of the 68 binding and relational cells of the 17
models, before and after). At a fixed depth, the span of the needed statements relates only weakly to
the share (mean $t=-0.17$; $|t|>1.96$ in 9 of the 68 cells, 7 of them negative, against 3.4 expected),
so most of the response follows $k$ rather than where the needed statements lie; across depths, span
and $k$ cannot be separated.

\paragraph{Wordings.} Pointer w0: ``Rules: $a$ goes to $b$. \ldots\ Start at $s$. Take $k$ steps. You
end at''; w1: ``Map: $a$ leads to $b$. \ldots\ From $s$, make $k$ moves. You finish at''.
Arithmetic w0: ``Rule: add $d$ and keep the remainder after dividing by 7. Start at $s$. Apply the rule
$k$ times. You end at''; w1: ``x = $s$. Repeat $k$ times: x = (x + $d$) mod 7. Now x =''.
Binding w0: ``Assignments: $a$ = $b$. \ldots\ Therefore $v$ =''; w1: ``Facts: Let $a$ be $b$.
\ldots\ The value of $v$ is''. Relational w0: ``Chart: The boss of $A$ is $B$. \ldots\ The top boss
above $V$ is''; w1: ``Chart: $A$ reports to $B$. \ldots\ $V$ ultimately reports to''.

\section{Accuracy and correctness}
\label{app:accuracy}

\begin{table}[h]
\centering\small
\caption{Mean strict accuracy over the 17 models (both wordings, all items; Qwen2.5-7B-Instruct under
SDPA). An item counts as strictly correct when both the prediction restricted to admissible digits and the
first generated digit are correct.}
\label{tab:acc}
\begin{tabular}{@{}lccccccc@{}}
\toprule
Family & Chance & $k=1$ & 2 & 3 & 4 & 5 & 6 \\
\midrule
Pointer chasing & 0.10 & 0.57 & 0.16 & 0.09 & 0.07 & 0.06 & -- \\
Modular arithmetic & 0.14 & 0.63 & 0.15 & 0.16 & 0.13 & 0.12 & 0.11 \\
Variable binding & 0.33 & 0.86 & 0.61 & 0.44 & 0.34 & 0.32 & 0.28 \\
Relational lookup & 0.33 & 0.76 & 0.45 & 0.25 & 0.27 & 0.23 & 0.27 \\
Numeral control & 0.14 & 0.41 & 0.30 & 0.17 & 0.15 & 0.14 & 0.03 \\
\bottomrule
\end{tabular}
\end{table}

\begin{figure}[H]\centering
\includegraphics[width=\textwidth]{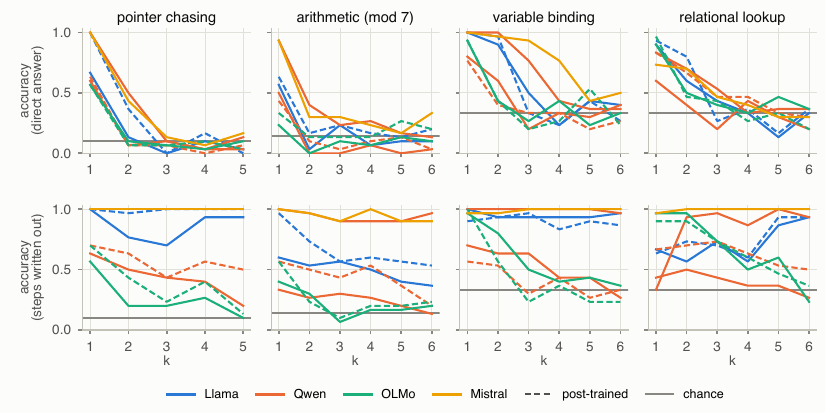}
\caption{\textbf{Accuracy on fresh items.} Accuracy by serial depth $k$ in the replication (\cref{app:replication}) for the eight rerun models, counting the answer each model writes. \emph{Top:} the model answers directly, as in
the main run. \emph{Bottom:} the worked examples and the model write out the intermediate states before
the answer. Color: family; dashed: post-trained models; gray line: chance, one over the number of admissible answers (the worked examples rule out some answers; \cref{app:replication}).}
\label{fig:acc}
\end{figure}

\paragraph{Accuracy across depths.} Theoretical results on fixed-depth transformers
\citep{sanford2024,merrill2024} and studies of multi-hop reasoning \citep{press2023,biran2024,yang2024}
lead us to expect accuracy to decrease with the number of steps composed within one forward pass, and it
does so in all four families (\cref{tab:acc}; per model on fresh items in \cref{fig:acc}). At $k=1$, every model answers
above chance in arithmetic, binding and relational lookup, and 16 of the 17 do so in pointer chasing
(one-sided $t$-test over the 42 instances, each averaged over the two wordings). Pointer chasing has the
lowest single-step accuracy of the four families (mean 0.57). At $k=2$, the number of models above chance
is 4 in pointer chasing, 3 in arithmetic, 13 in binding and 7 in relational lookup, and at $k\ge3$ mean
accuracy is 0.07, 0.13, 0.34 and 0.26, against chance levels of 0.10, 0.14, 0.33 and 0.33. The decrease
comes with composition rather than with the single step. In pointer chasing, the four models with a $k=1$
accuracy of at least 0.90 answer 0.30--0.54 of the $k=2$ items correctly, and in arithmetic the two such
models answer 0.38--0.40, whereas two independent steps at each model's $k=1$ accuracy, with a
chance-level guess otherwise, would give 0.84--1.00 and 0.92--0.96. In binding, eight of the nine models
with a $k=1$ accuracy of at least 0.90 have an accuracy below 0.65 at $k=3$. Accuracy averaged over the
four serial families increases with parameter count among the 11 base models (Spearman $\rho=0.94$ over all depths, 0.93 at $k=1$ and 0.78 at $k=3$); at $k\ge3$ it lies between 0.15 and 0.24 in 16 of the 17 models and is 0.31 in
Mistral-Small-24B, against a mean chance level of 0.23. When the eight models of the replication first
write out the intermediate states, as in chain-of-thought prompting \citep{wei2022}, accuracy rises at
every $k\ge2$ in every family and, at $k\ge3$, in
every model, from 0.23 to 0.62 on average (\cref{fig:acc}, bottom; \cref{app:replication}).

\paragraph{Checks of the setup.} Besides re-deriving every answer from the prompt text
(\cref{sec:tasks}), we checked that prompt length in tokens is constant across the items of each run and
that every pointer table is a single 10-cycle, so that the fewest lookups needed equal $k$. At the answer
position, the digit tokens receive on average 0.99 of the next-token probability, and the first generated
character is a digit in 99.7\% of answers; all but three of the other answers come from Qwen2.5-7B under
eager attention (\cref{app:robust}). The prediction restricted to admissible digits and the first
generated digit agree on 95\% of items, and scoring either one alone gives an overall accuracy of 0.30 and
0.31, respectively, against 0.29 under strict scoring. When the gold digit ties with another digit for
the highest logit (4.6\% of items at bf16 precision), we count the item as incorrect; splitting such ties
evenly would raise accuracy under the restricted prediction from 0.30 to 0.32.

\paragraph{Errors.} Incorrect answers are structured: in pointer chasing and arithmetic, they are
intermediate states of the chain more often than chance would give, much as models often resolve only
the first hop of two-hop queries \citep{biran2024,yang2024}. At $k\ge2$, 47\% of incorrect
pointer answers are an intermediate state, against 28\% if incorrect answers were spread evenly over the
other digits ($n=5{,}174$), and 64\% of incorrect arithmetic answers are an intermediate value
$(s+jd)\bmod7$ with $1\le j<k$, against 50\% under the same baseline ($n=6{,}167$). Incorrect answers
repeat the answer of a worked example within five points of the rate expected by chance (34\%, 36\%,
32\% and 12\% at $k\ge2$ in the four families, against 30\%, 41\%, 31\% and 13\%), so copying from the
worked examples does not account for them.

\paragraph{Correct and incorrect answers.} At a given $k$, head activity at the answer position is
nearly the same for correct and incorrect answers, so we analyze all items. Cohen's $d$ of the top-10\%
share between correct and incorrect answers at the same depth (positive when correct answers have the
larger share) averages $+0.014$ over the 4{,}024 layers of the 128 serial cells with at least eight correct and eight incorrect items at some depth, just outside the range
holding 95\% of a null that shuffles correctness within depth ($\pm0.010$). Individual layers differ more
often than under the null (15\% at $p<0.05$, against 5\%), but with signs that vary across layers and
cells. Two of the 23 family $\times$ depth comparisons survive Benjamini--Hochberg correction, relational
lookup at $k=1$ and binding at $k=4$, and in both, correct answers are slightly more concentrated in the
back half ($d=+0.08$ and $+0.07$). The layer profile does not, on average, predict correctness better
than the prompt does: with cross-validation that keeps each prompt body in one fold, the median AUC over
the 125 of 136 serial cells with enough correct and incorrect items is 0.53 from the top-10\% shares and
0.61 from prompt features such as the gold digit, and adding the shares to the prompt features lowers the
median AUC by 0.01. The model differences also hold among incorrect answers: recomputed from the items
answered incorrectly in both wordings, the signatures correlate 0.79 with those from all items, slightly
below random subsets of the same size (0.80--0.84), and model identity explains 27\% of the non-shared
variation ($p=0.003$), about as much as in those subsets (28\%).

Correctness is associated more with how strongly the share changes with $k$. In back-half layers, the
top-10\% share falls by 0.10 percentage points per step over all items (\cref{app:shared}), and 0.06
percentage points per step faster among correct than among incorrect answers ($p=0.001$). Most of this
difference comes from the $k=1$ items (0.02 without them, $p=0.06$), except in binding, whose back-half
spreading comes mostly from correct answers (0.25 against 0.02 percentage points per step; 0.09 without
the $k=1$ items, $p=0.001$). For the row ``Adjusted for entropy and correctness'' of \cref{tab:robust},
we add the entropy of the answer distribution over the admissible digits and strict correctness as
item-level covariates to the regression on $k$ and use the $t$ statistic of the $k$ coefficient in place
of $z$. The resulting signatures correlate 0.86 with the primary ones. The same item-level regression gives 0.98
without covariates, 0.96 with correctness alone and 0.89 with entropy alone, so the lower correlation
comes mostly from the entropy covariate rather than from correctness.

\paragraph{Prompt variants.} Before the main run, we compared its settings (three worked examples, three
chains and a 10-state pointer) with variants on a separate calibration split. We ran this comparison in
12 of the 17 models (all except the two Mistral models and the three Llama and Qwen2.5 Instruct models)
and in Llama-3.1-70B, adding Qwen2.5-72B in the two 10-state pointer settings, and scored 18 items per
depth and wording by the prediction restricted to admissible digits. Eight worked examples instead of
three, still at depths 1--3, change accuracy at $k=1$ by $+0.07$ in arithmetic, $-0.12$ in pointer
chasing and $-0.13$ in the numeral control, and at $k\ge2$ by $-0.02$, $+0.03$ and $+0.01$. Two chains
instead of three raise accuracy at $k\ge2$ by 0.19 in binding and 0.18 in relational lookup, mostly
because chance rises from 1/3 to 1/2: accuracy minus chance rises by 0.02 in both (95\% bootstrap
intervals include zero), and the gain as a fraction of the distance from chance to perfect accuracy is
0.08 ($[0.04, 0.14]$) and 0.06 ($[-0.02, 0.14]$). A 7-state pointer with $k\le3$ and eight worked
examples matches the 10-state pointer at $k=2$ and 3 in accuracy minus chance (difference $-0.01$ against
eight worked examples and $+0.004$ against three; both intervals include zero). In the replication, eight
worked examples that cover every depth and never show the query's answer give nearly the same accuracy at $k\ge3$ as the main run (0.23 against 0.22; \cref{app:replication}), and the per-family differences at $k=2$ and at $k\ge3$
lie between $-0.06$ and $+0.07$. Within one forward pass, prompt variants thus change accuracy at $k=1$
by up to 0.17 (arithmetic on fresh items) and accuracy at $k\ge2$ by at most 0.07, after allowing for the
change in chance where a variant changes it, whereas writing out the steps raises accuracy at $k\ge3$ by
0.38. Accuracy at larger $k$ is therefore set mainly by composition within one forward pass rather than by
the prompt settings.

\section{Replication on fresh items}
\label{app:replication}

\paragraph{Design.} We reran the four serial families in eight of the models: Llama-3.1-8B and
Llama-3.1-8B-Instruct, Qwen2.5-7B and Qwen2.5-7B-Instruct, Qwen2.5-14B, OLMo-2-1124-7B and
OLMo-2-1124-7B-Instruct, and Mistral-Small-24B, which form five lineages from all four families. The
items are newly generated, 30 per depth, so a cell holds 150 items in pointer chasing and 180 in the
other families; as in the main run, arithmetic repeats problems within a depth (20--26 distinct problems
per depth, each with different worked examples). All prompts use wording w0. Each prompt has eight
worked examples in random order, drawn independently of one another (each with its own table in pointer
chasing and its own chains in binding and relational lookup), whose depths cover every $k$ of the
family. No worked example has the query's answer, whereas the main run balanced worked-example answers
over the admissible set. The worked examples therefore rule out some answers, to a similar extent at
every depth: choosing at random among the answers that no worked example shows (and that differ from
the written step count in pointer chasing and arithmetic) would score 24--27\% in pointer chasing,
48--57\% in arithmetic and 63--73\% in binding and relational lookup at each $k$. Binding and
relational lookup interleave three chains of six statements (18 in all) in shuffled order and add a
question line before the query (``Question: what is v? Therefore v ='' and ``Question: who is the top
boss above V? The top boss above V is''). Every item has its own prompt body, so the permutation null
shuffles $k$ across items in all four families. With six statements per chain, the entity queried at
$k=6$ is the last of its chain and appears once in the body rather than twice; without the $k=6$
binding and relational items, the agreement with the main run, measured by one correlation over the signatures of all eight models (pooled $r$), barely changes (0.75, against 0.76). There is no numeral control. All eight models ran in bf16 with eager
attention, including Qwen2.5-7B-Instruct, which the main run analyzes under SDPA. The OLMo-2 exclusion
(\cref{sec:measure}) removes 28 of 180 arithmetic items, 7 binding items and 1 relational item for the
base model, and 42 of 180 binding, 5 arithmetic, 4 relational and 2 pointer items for the Instruct model;
no other model has a flagged item. Answers are balanced within each depth and, in pointer chasing and
arithmetic, never equal the written step count. An independent solver re-derived the answers of all
690 prompts and their 5{,}520 worked examples from the prompt text, and the per-head writes sum to the
output projection's output with a relative error of at most $8\times10^{-3}$. We also ran every item in
a second mode, in which the worked examples write out their intermediate states (``Path: 4 -> 9 -> 1 ->
7. You end at 7.'') and the model writes its own steps before answering. We use this mode only for
accuracy.

\paragraph{Accuracy.} Counting the answer each model writes and averaging over the four families (\cref{fig:acc}), the
eight models answer 36\% of the items correctly and 23\% at $k\ge3$, against 36\% and 22\% in the main
run on the same models (in the replication, strict and written-answer accuracy agree to two decimals;
the main-run values are strict accuracy). Arithmetic at $k=1$
is harder on the fresh items (0.57 against 0.74, lower in all eight models), and 37\% of pointer
predictions at $k=2$ are again the state one step from the start (39\% in the main run on these models).
When the models first write out the intermediate states, accuracy at $k\ge3$ rises to 0.62 (95\%
bootstrap interval across models 0.43--0.80). It rises in every model (McNemar $p\le0.001$), to between
0.31 for OLMo-2-1124-7B-Instruct and 0.98 for Mistral-Small-24B.

\begin{figure}[H]\centering
\includegraphics[width=0.545\textwidth]{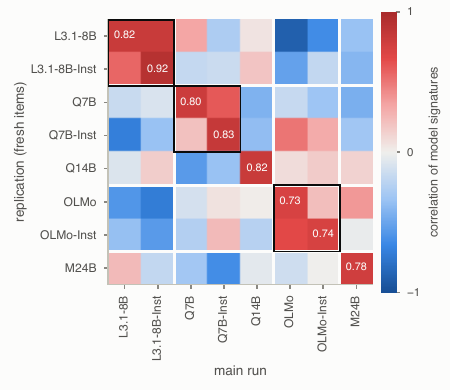}
\caption{\textbf{Signatures on fresh items against the main run.} Correlation between each model's
signature in the replication (rows) and in the main run (columns, both wordings), with signatures
centered on the eight models. Black squares: lineages; numbers: a model's correlation with its own
main-run signature (mean 0.80). The off-diagonal entries average $-0.11$, about $-1/7$ of the diagonal
mean, as centering implies. Llama-3.1-8B resembles its own and its Instruct model's main-run signature
equally (both 0.82). L: Llama; Q: Qwen2.5; M: Mistral.}
\label{fig:replication}
\end{figure}

\paragraph{Signatures.} \Cref{tab:replication} compares the two runs on these eight models, with
signatures centered on them. Within the replication, model identity explains 29\% of the non-shared
variation (19\% by chance, $p=0.003$), signatures estimated from two tasks predict those from the other
two (split-half $r=0.31$, $p=0.005$), and a model on two different tasks resembles itself more ($r=0.26$)
than different models on the same task (0.17). On the same models, the main run gives 30\% (18\% by
chance), $r=0.36$, and 0.32 against 0.30. Across runs, the shared profiles correlate 0.93, and
single-cell profiles agree at $r=0.69$ on average (positive in all 32 cells), against 0.21 between
different models on the same task. A model's replication signature correlates $r=0.80$ on average with
its main-run signature (0.73--0.92; \cref{fig:replication}; exact label permutation over all $8!$
assignments, $p=2.5\times10^{-5}$). Its correlations with the main-run signatures of the other models
average $-0.11$, about the $-0.80/7$ that centering implies for this own-signature mean. Every model
resembles its own main-run signature more than those of the other models on average, and seven of the
eight resemble their own most; Llama-3.1-8B resembles its own and its Instruct model's main-run
signature equally (both 0.82). Lineage partners again have similar signatures (mean $r=0.64$) and
models from different pretraining runs do not ($-0.23$, against $-0.14$ expected from centering), but the
three lineages change order: OLMo-2 is the most similar pair in the main run (0.69) and the least similar
in the replication (0.50). Qwen2.5-14B, whose signature does not replicate across tasks in the main run,
does so in the replication (split-half $r=0.41$), and its signature again differs from that of
Qwen2.5-7B ($r=-0.49$, against $-0.45$ in the main run on these models).

\paragraph{Items and wording.} The replication keeps wording w0 and changes the items, the worked
examples and the format of the chain-lookup prompts. Its signatures agree more closely with the main-run
w0 signatures (mean own $r=0.83$, pooled 0.79, all eight models most similar to themselves) than with
the w1 signatures (0.65, pooled 0.63), and at least as closely as the main-run w0 and w1 signatures
agree with each other on the same items (0.73, pooled 0.76). A change of wording thus moves a model's
signature at least as much as a new sample of items and worked examples does.

\paragraph{Kernel, model subset and exclusion.} With Qwen2.5-7B-Instruct taken from the main run's
eager capture, the pooled agreement is 0.74 and all eight models resemble their own main-run signature
most. Without the Qwen2.5-7B pair, the pooled agreement is 0.75 (exact $p=1/720$), and four of the six
models resemble their own main-run signature most, while Llama-3.1-8B and OLMo-2-1124-7B-Instruct
resemble that of their lineage partner. Without the OLMo-2 exclusion, the replication signatures
correlate 0.99 with those reported here.

\paragraph{Features described in the main text.} With one wording, no rerun model deviates from the average rerun model's back-half response in the same direction in all four tasks. The responses are weaker on fresh items: the mean
back-half $z$ is $-0.72$, against $-1.71$ in the main run on the same models and $-1.09$ when the main
run is subsampled to the replication's design (one wording, 30 items per depth, $k$ shuffled across
items; mean over random subsamples), so the design alone accounts for about 60\% of the difference.
\Cref{tab:replication-features} therefore compares the two runs mainly by sign, count of models and
correlation. In back-half layers, the top-10\% share falls by 0.09 percentage points per step (95\% CI
across models 0.04--0.15). The spreading at 50--70\% of depth, which the first principal component
measures, replicates across models ($r=0.82$, exact $p=0.005$), and the replication signatures projected
onto the main run's first component keep the models' positions ($r=0.84$, $p=0.007$). The back-half
mean replicates less well ($r=0.68$, $p=0.05$), although both of its parts replicate (50--80\%:
$r=0.88$, $p=0.005$; 80--100\%: $r=0.85$, $p=0.02$). Across models the two parts are negatively
correlated in both runs ($r=-0.27$ and $-0.50$), so their mean varies less between models than either
part does. Qwen2.5-7B's mean over 50--80\% is again the only positive one among the eight models
($+0.28$, against $+0.48$ in the main run). The locations also recur: the decile of strongest spreading
is the same in seven of the eight models (2.5 expected under label permutation, $p=0.004$), and 11 of
the 14 layers that spread consistently in the replication also do so in the main run (1.6 expected if
they were placed uniformly, 2.9 if placed uniformly within the back half). At 40--50\% of depth, the
OLMo-2 base model is again positive in all four tasks ($z=+3.5$, $+4.4$, $+0.7$ and $+1.0$), whereas
Mistral-Small-24B is negative in pointer chasing and arithmetic ($-2.0$ and $-1.6$). The final-layer
pattern that we found while exploring the main run holds in all eight models: the final layer spreads in
the three Qwen2.5 models and concentrates in the other five (label permutation $p=1/56$). On fresh items,
each post-trained model spreads more in the back half than its base model, including
OLMo-2-1124-7B-Instruct, whose change is close to zero in the main run, so the size of the change within
a lineage differs between runs. One main-run pattern does not recur: model deviations agree no more
between binding and relational lookup than between pointer chasing and arithmetic ($r=0.25$ each,
against 0.52 and 0.17 in the main run). The concentrating band at 40--50\% of depth is not significant
on these eight models on average in either run, so the replication tests it only model by model.

\begin{table}[H]
\centering\small
\caption{Main run and replication on the eight models run in both, with signatures centered on these
models. \emph{Top:} statistics within each run. Permutation $p$ uses 5{,}000 label shuffles for the model
share and 2{,}000 for the split-half $r$; the lineage test is exact over all $8!$ relabelings, and 0.002
is its smallest attainable $p$. Same model / different models: mean correlation of full task profiles.
Within a run, centering alone makes the mean correlation between different models' signatures
$-1/7\approx-0.14$. \emph{Bottom:} correlation of each model's replication signature with its main-run
signature, estimated from both wordings, from w0 alone and from w1 alone; the last column correlates each
model's main-run w0 and w1 signatures, which come from the same items. Across runs, centering makes a
model's correlations with the other models average about $-1/7$ of its own $r$. Pooled $r$: one correlation over the signatures of all eight models. Every mean own $r$ and pooled $r$ in the bottom part has exact permutation $p=2.5\times10^{-5}$.}
\label{tab:replication}
\begin{tabular}{@{}lcc@{}}
\toprule
Within each run & Main run & Replication \\
\midrule
Model share of non-shared variance (chance) & 0.30 (0.18) & 0.29 (0.19) \\
\quad permutation $p$ & $<0.001$ & 0.003 \\
Split-half $r$ across tasks & 0.36 & 0.31 \\
\quad permutation $p$ & $<0.001$ & 0.005 \\
Same model on different tasks / different models on the same task & 0.32 / 0.30 & 0.26 / 0.17 \\
Lineages: Llama-3.1-8B, Qwen2.5-7B, OLMo-2-1124-7B & 0.61, 0.39, 0.69 & 0.80, 0.62, 0.50 \\
\quad mean (exact $p$) & 0.56 (0.002) & 0.64 (0.002) \\
Different pretraining runs, mean & $-0.22$ & $-0.23$ \\
\bottomrule
\end{tabular}

\vspace{0.9em}
\begin{tabular}{@{}lcccc@{}}
\toprule
 & \multicolumn{3}{c}{Replication against the main run} & Main run \\
\cmidrule(lr){2-4}\cmidrule(l){5-5}
Own-signature $r$ & both wordings & w0 & w1 & w0 against w1 \\
\midrule
Llama-3.1-8B & 0.82 & 0.85 & 0.64 & 0.67 \\
Llama-3.1-8B-Instruct & 0.92 & 0.86 & 0.74 & 0.49 \\
Qwen2.5-7B & 0.80 & 0.79 & 0.74 & 0.86 \\
Qwen2.5-7B-Instruct & 0.83 & 0.90 & 0.74 & 0.93 \\
Qwen2.5-14B & 0.82 & 0.75 & 0.65 & 0.65 \\
OLMo-2-1124-7B & 0.73 & 0.86 & 0.37 & 0.63 \\
OLMo-2-1124-7B-Instruct & 0.74 & 0.84 & 0.57 & 0.69 \\
Mistral-Small-24B & 0.78 & 0.82 & 0.76 & 0.91 \\
\midrule
Mean own $r$ & 0.80 & 0.83 & 0.65 & 0.73 \\
Mean $r$ with other models & $-0.11$ & $-0.12$ & $-0.09$ & $-0.10$ \\
Own signature most similar & 7 of 8 & 8 of 8 & 7 of 8 & 6 of 8 \\
Pooled $r$ & 0.76 & 0.79 & 0.63 & 0.76 \\
\bottomrule
\end{tabular}
\end{table}

\begin{table}[H]
\centering\small
\caption{Features described in the main text, on the eight models run in both. The replication's $z$
values are about half as large (see text), so the columns are best compared by sign, count of models and
correlation. Intervals are 95\% $t$-intervals across the five lineages. Consistent layers: significant in
the same direction in at least three tasks and in the opposite direction in none (as in
\cref{tab:layers}); shared: also consistent in the main run. Post-training: change in the back-half mean $z$ over layers from the base to the Instruct model, averaged over the four tasks; the main-run Qwen2.5-7B pair mixes attention kernels
($-0.82$ with both models under eager attention). Across-run $r$: correlation over the eight models,
with exact permutation $p$; for the shared profile $\mu$, over the ten deciles.}
\label{tab:replication-features}
\begin{tabular}{@{}lll@{}}
\toprule
Feature & Main run & Replication \\
\midrule
Back half spreads & 8 of 8 & 8 of 8 \\
Back half spreads more than the front half & 8 of 8 & 8 of 8 \\
Mean $z$ at 50--80\% of depth, lineage-weighted & $-2.22$ $[-3.52, -0.92]$ & $-1.02$ $[-1.70, -0.34]$ \\
Mean $z$ at 40--50\% of depth, lineage-weighted & $+0.75$ $[-0.91, 2.42]$ & $+0.20$ $[-1.29, 1.69]$ \\
40--50\% decile above the 50--60\% decile & 8 of 8 & 6 of 8 \\
Consistently spreading layers & 33 & 14 (11 shared) \\
Final layer: Qwen2.5 spreads, others concentrate & 8 of 8 & 8 of 8 ($p=1/56$) \\
Agreement of model deviations: binding and relational & 0.52 & 0.25 \\
\quad pointer chasing and arithmetic & 0.17 & 0.25 \\
\quad across the two task types & 0.17 & 0.14 \\
Post-training change: Llama-3.1-8B & $-1.49$ & $-0.73$ \\
\quad Qwen2.5-7B & $-1.57$ & $-0.46$ \\
\quad OLMo-2-1124-7B & $-0.05$ & $-0.96$ \\
\midrule
Same decile of strongest spreading in both runs & \multicolumn{2}{l}{7 of 8 (2.5 expected, $p=0.004$)} \\
Across-run $r$: mean $z$ at 50--70\% of depth & \multicolumn{2}{l}{0.82 ($p=0.005$)} \\
Across-run $r$: projection on the main run's PC1 & \multicolumn{2}{l}{0.84 ($p=0.007$)} \\
Across-run $r$: mean $z$ at 50--80\% / 80--100\% of depth & \multicolumn{2}{l}{0.88 ($p=0.005$) / 0.85 ($p=0.02$)} \\
Across-run $r$: back-half mean $z$ & \multicolumn{2}{l}{0.68 ($p=0.05$)} \\
Across-run $r$: shared profile $\mu$ & \multicolumn{2}{l}{0.93} \\
\bottomrule
\end{tabular}
\end{table}

\section{A matched pair at 70B scale}
\label{app:large}

Llama-3.1-70B and Qwen2.5-72B have the same number of layers (80) and heads per layer (64), so the
comparison of this pair is free of the head-count confound of the main sample (\cref{sec:track}). We take
their head activity from the unablated forward passes of the earlier ablation on these two models
(\cref{app:ablation}), which cover every test item of wording w0 (42 per depth; 35 in pointer chasing)
under SDPA attention. On identical items, such captures reproduce the main run's per-head activity (for
Llama-3.2-3B in relational lookup, correlation 0.99995). We compute the top-10\% share, its response $z$ to
$k$ and the back-half test as for \cref{fig:contrast}, but from one wording.

\begin{figure}[H]\centering
\includegraphics[width=\textwidth]{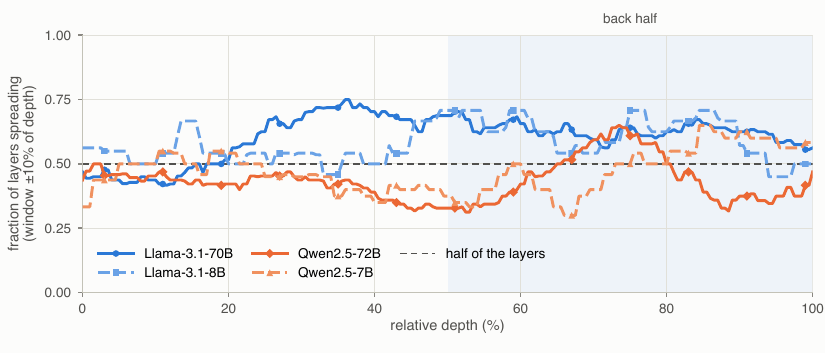}
\caption{\textbf{A majority of the Llama models' layers spread through most of the back half; in the
Qwen2.5 models, only in parts of it.} At each relative depth, the fraction of layers whose top-10\% head
share falls with $k$ ($z<0$), among the layers within $\pm10\%$ of that depth in the four serial tasks
(pooled). Llama-3.1-70B and Qwen2.5-72B have the same number of layers and heads (data of
\cref{app:large}); Llama-3.1-8B and Qwen2.5-7B come from the main run. The fraction exceeds one half at
100\% and 87\% of back-half positions in Llama-3.1-70B and 8B, against 28\% and 43\% in Qwen2.5-72B and
7B. Among the models not drawn, it does so at 73\% and 75\% in Llama-3.2-1B and 3B, at 0--26\% in
Qwen2.5-0.5B, 1.5B and 3B, and at 100\% in Qwen2.5-14B, which spreads like the Llama models.}
\label{fig:window}
\end{figure}

\Cref{tab:large} gives the results. In pointer chasing, arithmetic and relational lookup, Llama-3.1-70B
spreads strongly in the back half, as the smaller Llama models do, whereas the back half of Qwen2.5-72B
changes little on net and significantly only in pointer chasing, where it spreads weakly. Qwen2.5-72B does
spread in narrow bands, at 60--80\% of depth in pointer chasing and at 70--80\% in binding, but it
concentrates at 80--90\% of depth in all four tasks, where Llama-3.1-70B spreads in three, and it
concentrates strongly just before mid-depth in pointer chasing and arithmetic ($z=+6.2$ and $+6.9$ at
40--50\%). In binding, the pattern differs: Llama-3.1-70B concentrates in the back half, and Qwen2.5-72B
does not change significantly. The level of concentration differs as well: the back-half top-10\% share
is 0.52--0.58 in Llama-3.1-70B and 0.34--0.38 in Qwen2.5-72B across the four tasks. Pooled over the
four tasks, a majority of Llama-3.1-70B's layers spread at every point of its back half, whereas
Qwen2.5-72B has such a majority only in parts of it (\cref{fig:window}).

\begin{table}[h]
\centering\small
\caption{Response of the top-10\% head share to $k$ in a matched pair (80 layers, 64 heads per layer;
wording w0, SDPA). Back half: mean $z$ over layers at $\ge$50\% of depth, with its two-sided permutation
$p$; 80--90\%: mean $z$ in that decile. Negative: spreads; positive: concentrates.}
\label{tab:large}
\begin{tabular}{@{}lcccc@{}}
\toprule
& \multicolumn{2}{c}{Llama-3.1-70B} & \multicolumn{2}{c}{Qwen2.5-72B} \\
\cmidrule(lr){2-3}\cmidrule(l){4-5}
Task & back half ($p$) & 80--90\% & back half ($p$) & 80--90\% \\
\midrule
Pointer chasing & $-3.48$ ($<0.001$) & $-5.3$ & $-0.79$ (0.003) & $+3.2$ \\
Modular arithmetic & $-2.94$ ($<0.001$) & $-6.7$ & $-0.18$ (0.50) & $+1.5$ \\
Variable binding & $+1.37$ (0.002) & $+1.3$ & $-0.43$ (0.07) & $+4.5$ \\
Relational lookup & $-1.83$ ($<0.001$) & $-2.4$ & $-0.06$ (0.80) & $+2.2$ \\
\bottomrule
\end{tabular}
\end{table}

\section{The shared component in detail}
\label{app:shared}

\paragraph{Depth profile.} Weighting each lineage once, the mean $z$ values over tasks for the ten
deciles, from 0--10\% to 90--100\%, are $+0.36$, $+0.44$, $-0.06$, $-0.06$, $+1.19$, $-1.21$, $-1.86$,
$-0.97$, $-0.76$ and $-0.62$, and the 95\% confidence interval across lineages excludes zero at
40--80\%. Per task and averaged over the 17 models, the back half spreads clearly in pointer chasing, binding and relational lookup,
with mean back-half $z$ of $-1.22$, $-2.48$ and $-1.21$ and negative values in 16, 15 and 14 of 17
models. It spreads only weakly in arithmetic ($-0.34$, 95\% CI across models $[-0.88, 0.25]$), where
the back-half $z$ is negative in 11 of 17 models. The concentrating band at 40--50\% is
significant in arithmetic ($+3.48$), positive in pointer chasing ($+1.19$, 11 of 17 models) and
relational lookup ($+0.61$), and absent in binding ($-0.63$), so it comes mainly from arithmetic. The numeral control has a
concentrating band of its own, one decile earlier (30--40\%), which we found while exploring.

\paragraph{Tasks.} Differences in $z$ between tasks are only partly comparable, because binding and
relational lookup use a permutation null within prompt bodies, which makes their $|z|$ about 1.6 times
larger than under the null used for pointer chasing and arithmetic. The replication (\cref{app:replication})
uses one null for all four tasks. There, only binding spreads clearly in the back half (mean $z=-1.21$, 95\% CI across the eight models $[-2.08,-0.35]$), followed by pointer chasing ($-0.82$) and arithmetic ($-0.54$), and relational lookup spreads least ($-0.30$). In the main run, the cleanest comparison is between
binding and relational lookup, which share their design and their null: within models, binding spreads
more in the back half than relational lookup in 16 of 17 models (10 of 11 lineages, and 14 or 15 of 17
under the other statistics; 6 of 8 on fresh items), and it is also the more accurate of the two in 14 of 17. Measured by the slope as a percentage of the layer's mean share, binding spreads most in 9 of 17 models and pointer chasing in 6, but neither leads in a majority
of lineages (4 and 5 of 11), and their lineage means overlap. Arithmetic is the weakest task by this measure in 11 of 17 models, with no significant back-half spreading on average; among the eight rerun models it is the weakest in five in the main run and in two on fresh items. 

\paragraph{Cells.} Of the 136 model $\times$ task $\times$ wording cells (Qwen2.5-7B-Instruct under
SDPA), the back half spreads in 80\% and more than the front half in 84\%, and its mean $z$ is
significantly negative in 88 cells and significantly positive in 12. The result holds for splits at
40\% and 60\% of depth, with SDPA attention, without sink heads and for every statistic (81\% and 86\%
under the top-25\% share, 85\% and 88\% under the effective number of heads). An item-level model with a depth-by-$k$ interaction
gives $-0.110$ percentage points per step per unit of relative depth (95\% CI $[-0.175,-0.046]$,
clustered by model), and $+0.033$ ($[-0.053, 0.119]$) in the numeral control.

\paragraph{Anatomy.} In back-half layers the top-10\% share falls by 0.10 percentage points per step on
average (95\% CI 0.06--0.13; about 0.45 points over the range of $k$). How the fall arises differs by
task: in binding and relational lookup most of it is a loss of share by the layer's usual strongest heads (84\% and 72\% of the fall summed over cells), whereas in pointer chasing and arithmetic it is almost entirely a
flattening of item-specific peaks. In back-half layers, 38\% of heads lose activity significantly with
$k$ and 21\% gain (2.5\% and 3.1\% when $k$ is permuted). Total layer output falls with $k$ in the back
half of 88\% of cells (84\% in the replication); adjusting each layer's slope for its log output leaves a median 83\% of the back-half effect per cell (59\% of the effect summed over cells). Additive-floor models (a constant floor under every head's output)
explain at most about a third of it, unless the floors are placed only on the weaker heads. Adding
answer entropy as an item-level covariate reduces the back-half slope by 45\% and the depth-by-$k$
interaction, which remains significant, by a third. In the replication, the same adjustment reduces the
back-half response by 38\%, and the shared spreading at 50--80\% of depth halves but remains (mean
item-level $t$ from $-1.04$ to $-0.54$, 95\% CI $[-1.00,-0.08]$).

\paragraph{Post-training.} The post-trained models' stronger back-half spreading comes mainly from
pointer chasing and binding, and in the numeral control the lineage-weighted change is $-2.36$, against
$-0.95$ in the serial tasks. In the one cell where both Qwen2.5-7B models ran with SDPA (pointer chasing, wording w0), their back-half difference reverses. Every post-trained model is more accurate at $k=1$ than its base (by 0.06 to
0.20), and its answer entropy rises more steeply with $k$. On identical items, the post-trained model's faster fall of the top-10\% share at 40--80\% of depth shrinks by almost half without the $k=1$ items and disappears on items that both models answer incorrectly.

\section{Robustness of the signatures}
\label{app:robust}

\paragraph{Attention kernel.} Under eager bf16 attention, Qwen2.5-7B and 7B-Instruct agree with SDPA on
only 52--55\% of their predictions and are less accurate; Qwen2.5-1.5B agrees on 78\%, and the other
models agree more often. On the one Qwen2.5-7B cell captured under both kernels (pointer chasing, wording w0), the per-layer profiles
correlate $r=0.85$, but the back-half response is half as strong under eager attention ($z=-1.55$ against
$-3.05$), whereas for Llama-3.1-8B and OLMo-2-7B the two kernels agree ($r\ge0.99$). Under eager
attention, 5.5\% of Qwen2.5-7B-Instruct's generated answers begin with a non-digit after any leading space, against none under SDPA, as do 2.9\% of Qwen2.5-7B's, and in the replication, which ran both Qwen2.5-7B models under eager
attention, about a fifth of their written step traces contain malformed elements such as an empty state
(20.7\% and 18.6\%, against at most 0.6\% in the other six models).

\paragraph{Variants.} We remove attention-sink heads (mean attention to the first token above 0.9) as a fixed set per cell.
The \mbox{top-50\%} share depends more on the middle of the distribution; it gives less reliable signatures
(split-half 0.29) that correlate only 0.42 with the primary ones, but 0.71 with those of the top-25\%
share and 0.77 with those of the effective number of heads. Wording matters more than item sampling:
signatures estimated from wording w0 alone and from w1 alone correlate $r=0.75$ (0.47--0.93 per model),
and single-cell profiles agree across wordings at $r=0.54$, against 0.84 across item halves. The
replication, which keeps w0, accordingly agrees more with w0 (pooled signature $r=0.79$) than with w1 (0.63;
\cref{app:replication}).

\paragraph{Deviations with the same sign in every task.} Under the top-10\% share, seven models deviate
from the average back-half response in the same direction in all four tasks, against 1.8 expected when
model labels are shuffled within each task ($p<0.001$), and under the share of the strongest head nine do
(2.1 expected, $p<0.001$). Under each of the top-25\% and top-50\% shares and the effective number of heads, three models do (2.0--2.2 expected, $p=0.34$--0.36): Qwen2.5-0.5B and 1.5B, which spread less than average under all three statistics, and a third that changes with the statistic, Qwen2.5-7B (less) under the top-25\% share, Qwen2.5-7B-Instruct
(more) under the top-50\% share and Llama-3.1-8B-Instruct (more) under the effective number of heads.

\paragraph{Family contrast in relational lookup.} The complete separation of the three Llama base
models from the four Qwen2.5 base models up to 7B (\cref{fig:contrast}; exact one-sided permutation
$p=1/35$, the smallest attainable) holds under the top-25\% and top-50\% shares and the effective number
of heads, but not under the share of the strongest head ($p=0.086$). Under these three statistics,
however, Qwen2.5-7B also spreads in the back half ($-0.95$, $-1.92$ and $-1.49$; permutation
$p\le0.004$), as do Qwen2.5-1.5B and 3B under the last two ($p\le0.01$). What holds under these three statistics is therefore that each of these Qwen2.5 models spreads less than every Llama model, not that it does not
spread. The relational separation holds with wording w0 alone but not with w1 alone ($p=0.057$), where
Llama-3.2-1B ($-0.49$) and Qwen2.5-1.5B ($-0.59$) overlap. With Qwen2.5-14B included, $p=0.089$ ($0.048$
with OLMo-2 on the Llama side). In binding, the separation holds only under the top-10\% share; in
arithmetic, it fails under the top-50\% share and the effective number of heads. On fresh items, the
three models of \cref{fig:contrast} that were rerun keep their relational order (OLMo-2 $-0.66$,
Llama-3.1-8B $-0.33$, Qwen2.5-7B $+0.39$), but none of the three responses is significant. In the main run, across the eight models of \cref{fig:contrast} and Qwen2.5-14B, heads per layer correlate with the relational back-half response (Spearman
$\rho=-0.85$), so head count and family remain confounded.

\begin{table}[H]
\centering\small
\caption{Model signatures under alternative statistics, controls and subsets. Model share: share of
the non-shared variance explained by model identity, with its label-permutation chance level in
parentheses (permutation $p<0.001$ in every row).
Split-half: correlation of signatures estimated from two tasks with those from the other two. $r$:
correlation of the signatures with the primary ones; for the SDPA variant, with those under eager attention on the same 14 models, and for the no-exclusion variant, with primary signatures computed the same way, from $z$ per wording averaged over the two wordings. Lineage / other: mean signature correlation within and
between lineages.}
\label{tab:robust}
\begin{tabular}{@{}lccccc@{}}
\toprule
Variant & Model share & Split-half & $r$ & Lineage & Other \\
\midrule
Top-10\% share (primary) & 0.34 (0.21) & 0.36 & 1.00 & 0.73 & $-0.11$ \\
Top-25\% share & 0.30 (0.20) & 0.31 & 0.84 & 0.64 & $-0.11$ \\
Top-50\% share & 0.29 (0.20) & 0.29 & 0.42 & 0.71 & $-0.11$ \\
Effective number of heads & 0.31 (0.20) & 0.34 & 0.84 & 0.70 & $-0.11$ \\
Share of the strongest head & 0.36 (0.23) & 0.36 & 0.79 & 0.65 & $-0.11$ \\
Attention-sink heads removed & 0.34 (0.21) & 0.36 & 0.95 & 0.78 & $-0.12$ \\
SDPA attention (14 models) & 0.34 (0.22) & 0.35 & 0.98 & 0.76 & $-0.11$ \\
No OLMo-2 item exclusion & 0.33 (0.21) & 0.33 & 0.99 & 0.73 & $-0.11$ \\
Five depth bins & 0.35 (0.21) & 0.41 & -- & 0.79 & $-0.11$ \\
20-point depth interpolation & 0.36 (0.23) & 0.35 & -- & 0.73 & $-0.11$ \\
Base models only (11) & 0.33 (0.22) & 0.32 & -- & -- & -- \\
Models with $H=32$ only (9) & 0.30 (0.18) & 0.39 & -- & -- & -- \\
Adjusted for answer entropy & 0.30 (0.21) & 0.28 & 0.89 & 0.78 & $-0.12$ \\
Adjusted for entropy and correctness & 0.30 (0.21) & 0.27 & 0.86 & 0.77 & $-0.11$ \\
\bottomrule
\end{tabular}
\end{table}

\section{Ablation of the most active heads}
\label{app:ablation}

\paragraph{Design.} We ran all 17 models on the four serial families (wording w0, test split) in bf16
with SDPA attention; activity in the main analysis comes from eager attention (\cref{sec:measure}).
We split the items of each family in half at every depth (126 items per half, 105 in pointer chasing). On the
selection half we record each head's activity at the answer position, rank the heads of each layer by
their mean activity and store each head's mean input to the output projection. On the measurement
half we ablate, at the answer position only, the $q=\lceil 0.1H\rceil$ heads with the largest mean
activity in every layer of a set. Mean-ablation \citep{wang2023ioi} replaces a head's slice of the
output-projection input by its selection-half mean, which keeps the head's average output and removes
its item-specific part; zero-ablation removes both. The sets are fixed by depth, whatever the layers'
response to $k$: the 40--80\% window and five bands of 20\%, each ablated as a whole. The baseline
ablates $q$ heads per layer drawn at random from those between the 25th and 75th percentiles of mean activity,
excluding the $\max(q,3)$ most active (five draws for the window, two per band). The answer margin is the logit of the correct digit minus the mean logit of the other admissible answers. The drop is one minus the ratio of the ablated to the clean margin on items the model answers correctly, averaged within each depth with at
least five such items and then over depths; the excess drop subtracts the drop from random heads. A
model's share is its top-10\% share in the ablated layers, averaged over selection items. In pointer
chasing, Llama-3.2-1B and 3B and Qwen2.5-0.5B and 1.5B have no depth with five correct items, which
leaves 13 models. We fixed the design and the analysis before running the ablation. On the first item of every run, we check each ablated input against the intended one and each change in the projection's output against the ablated heads' writes; the cached computation reproduces a full forward pass within bf16 rounding.

\begin{figure}[H]
\centering
\includegraphics[width=\textwidth]{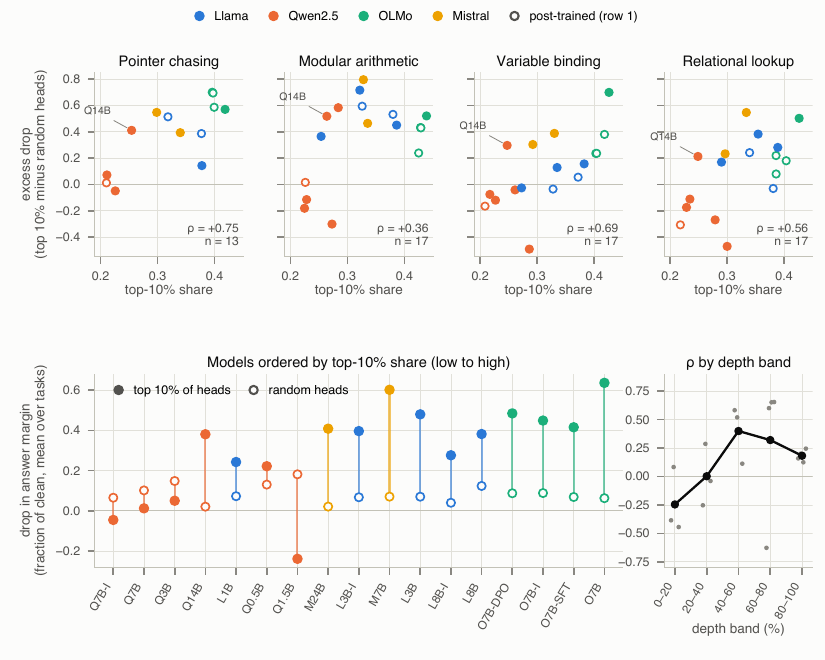}
\caption{\textbf{Models with more concentrated activity depend more on their most active heads.} The
top 10\% of heads per layer at 40--80\% of depth are mean-ablated at the answer position. Top: each
model's top-10\% share in those layers against its excess drop (loss of answer margin, as a fraction of
clean, from the top heads minus that from as many random mid-ranked heads; correctly answered items,
depth-balanced), per task, with Spearman $\rho$ across models. Open markers: post-trained models.
Bottom left: drops from the top heads (filled) and from random heads (open), averaged over tasks, with
models ordered by share. Bottom right: the same $\rho$ when each 20\% depth band is ablated on its own;
gray, each task; black, their mean. L: Llama; Q: Qwen2.5; O: OLMo-2; M: Mistral; -I: Instruct.}
\label{fig:ablation}
\end{figure}

\paragraph{Results.} Removing the top tenth of heads at 40--80\% of depth costs the median cell outside
Qwen2.5 46\% of the answer margin, against 6\% for random heads (\cref{fig:ablation}, bottom left). In
the Qwen2.5 models up to 7B, the median top-head drop is $-5\%$ (the margin grows) against 9\% for
random heads, and the excess drop is negative in 14 of their 18 cells, against 3 of the 42 cells of the
other families. Llama-3.1-8B and Qwen2.5-7B illustrate the contrast: in the four tasks their shares are
0.38--0.39 and 0.21--0.23 and their excess drops 0.14 to 0.45 and $-0.18$ to 0.07. Qwen2.5-14B behaves like
the other families (excess drop 0.21--0.52) despite a share of 0.25--0.26. The association between
share and excess drop holds under every variant we tried except those that remove the family
contrast (\cref{tab:ablation}), and the drop from random heads does not follow the share. The association appears when the bands at 40--60\% and 60--80\% of depth are ablated (mean $\rho=0.40$ and 0.32, $p=0.002$ and
0.007), not those in the first 40\% ($-0.25$ and 0.00). Under zero-ablation it weakens and vanishes in
relational lookup: zeroing the top heads of the Qwen2.5 models up to 7B removes a median 28\% of the
margin (up to 96\%), so these models need the constant part of their top heads' output but not its
item-specific part. A model's share varies little across tasks (mean range 0.02, against 0.21 across
models within a task), and share and excess drop do not correlate across a model's tasks (mean
$\rho=-0.05$); with so little variation this does not test the relation. The excess drop, averaged over
tasks, also follows the 50--70\% response of \cref{sec:track} ($\rho=-0.51$, $p=0.04$), again through
Qwen2.5 ($-0.06$ without it).

\paragraph{Two larger models.} An earlier ablation on Llama-3.1-70B and Qwen2.5-72B points the same way, though clearly only in arithmetic. It ablated the one or three most active heads per layer in those layers of the 40--80\% window whose top-10\% share changes significantly with $k$, rather than in fixed bands. In layers 32--63, Llama-3.1-70B's top-10\%
heads hold 49--56\% of summed head activity, against 34--37\% in Qwen2.5-72B. Mean-ablating three heads
per layer removes 34\% of Llama's answer margin in arithmetic, 13\% in pointer chasing and 10\% in
binding, against at most 9\% of Qwen's, while random heads remove about 1\% or less in both. In relational lookup, ablating Llama's top heads
increases its margin by 23\%, and Qwen2.5-72B has no layer to ablate.

\begin{table}[H]
\centering\small
\caption{Association between a model's top-10\% share and its excess drop under alternative
choices. Per task: Spearman $\rho$ across models (in parentheses, the number of models where it differs from the row's total, which is 17 unless the row gives another; pointer chasing has 13 in the primary analysis). Mean: average over tasks, with a one-sided
permutation $p$ that shuffles models within each task (10{,}000 permutations).}
\label{tab:ablation}
\setlength{\tabcolsep}{4pt}
\begin{tabular}{@{}lcccccc@{}}
\toprule
Variant & Pointer & Arithmetic & Binding & Relational & Mean & $p$ \\
\midrule
Primary & 0.75 & 0.36 & 0.69 & 0.56 & 0.59 & $<0.001$ \\
Share minus its floor $q/H$ & 0.77 & 0.40 & 0.76 & 0.64 & 0.64 & $<0.001$ \\
Partial on heads per layer & 0.67 & 0.31 & 0.63 & 0.47 & 0.52 & $<0.001$ \\
Models with $H\ge24$ (14) & 0.71 (12) & 0.19 & 0.61 & 0.42 & 0.48 & $<0.001$ \\
Base models (11) & 0.64 (7) & 0.55 & 0.64 & 0.67 & 0.63 & $<0.001$ \\
Cells with $\ge$15 correct items & 0.76 (8) & 0.43 (13) & 0.69 & 0.56 & 0.61 & $<0.001$ \\
Without OLMo-2 (13) & 0.42 (9) & 0.58 & 0.58 & 0.62 & 0.55 & $<0.001$ \\
Excess drop in logits & 0.57 & 0.54 & 0.73 & 0.42 & 0.56 & $<0.001$ \\
Top-head drop, no baseline subtracted & 0.80 & 0.39 & 0.77 & 0.56 & 0.63 & $<0.001$ \\
Zero-ablation drop & 0.61 & 0.38 & 0.69 & $-0.08$ & 0.40 & $<0.001$ \\
\addlinespace
Without Qwen2.5 (11) & 0.50 (9) & $-0.35$ & 0.55 & 0.04 & 0.18 & 0.14 \\
Within family (family means removed) & $-0.15$ & 0.30 & 0.35 & 0.00 & 0.13 & 0.17 \\
Random-head drop (control) & 0.21 & $-0.02$ & $-0.05$ & $-0.17$ & $-0.01$ & 0.52 \\
\bottomrule
\end{tabular}
\end{table}

\section{Consistently responding layers}
\label{app:layers}

\begin{table}[H]
\centering\small
\caption{Layers significant ($p<0.05$) in the same direction in at least three of the four tasks and
significant in the opposite direction in none (0-based layer index; relative depth in parentheses).
Over all models, 56 layers spread consistently against 39.5 expected when each task's pattern is
shifted to a random depth ($p=0.001$), and 20 concentrate against 11.0 ($p=0.003$).}
\label{tab:layers}
\setlength{\tabcolsep}{4pt}
\begin{tabular}{@{}lp{4.35cm}p{6.2cm}@{}}
\toprule
Model & Concentrate & Spread \\
\midrule
Llama-3.2-1B & -- & 4 (.27), 10 (.67) \\
Llama-3.2-3B & 12 (.44), 27 (1.0) & 19 (.70) \\
Llama-3.2-3B-Instruct & -- & 16 (.59), 18 (.67) \\
Llama-3.1-8B & 15 (.48), 31 (1.0) & 16 (.52), 17 (.55), 24 (.77), 26 (.84) \\
Llama-3.1-8B-Instruct & -- & 2, 13, 16, 17, 19, 24, 27, 30 (.06--.97) \\
Qwen2.5-0.5B & -- & 23 (1.0) \\
Qwen2.5-1.5B & 22 (.81) & -- \\
Qwen2.5-3B & 9 (.26), 19 (.54), 27 (.77) & 25 (.71) \\
Qwen2.5-7B & 20 (.74) & 4 (.15) \\
Qwen2.5-7B-Instruct & 12 (.44) & 17 (.63), 23 (.85) \\
Qwen2.5-14B & 15 (.32) & 29, 30, 35, 37, 47 (.62--1.0) \\
OLMo-2-7B & -- & 18, 19, 20, 21, 23 (.58--.74), 29 (.94) \\
OLMo-2-7B-SFT & -- & 3 (.10), 18, 19, 21 (.58--.68), 29 (.94) \\
OLMo-2-7B-DPO & 12 (.39) & 4 (.13), 18, 19, 21 (.58--.68) \\
OLMo-2-7B-Instruct & 12 (.39), 31 (1.0) & 18, 19, 21 (.58--.68) \\
Mistral-7B & 13 (.42), 14 (.45) & 4 (.13), 7 (.23), 18 (.58), 20 (.65), 26--28 (.84--.90) \\
Mistral-Small-24B & 3 (.08), 17 (.44), 19 (.49), 32 (.82) & 5 (.13), 12 (.31), 24 (.62), 27 (.69) \\
\bottomrule
\end{tabular}
\end{table}

\section{Additional figures}
\label{app:figs}

\begin{figure}[H]
\centering
\includegraphics[width=\textwidth]{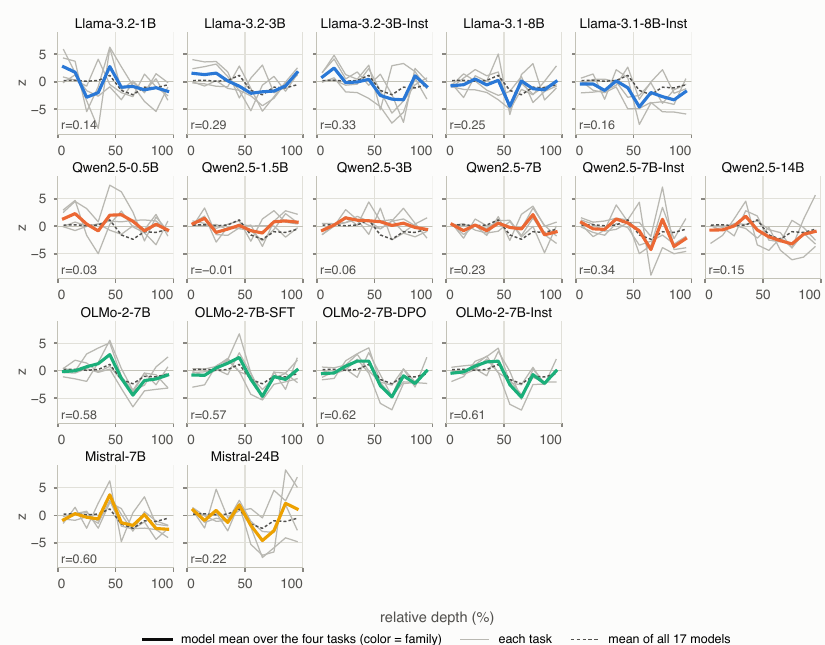}
\caption{\textbf{Each model's depth profile, per task.} Standardized slope $z$ of the
\mbox{top-10\%} head share on $k$, by relative depth (negative: the layer spreads activity as $k$
grows; positive: it concentrates). Gray: each of the four tasks; color: the model's mean over tasks;
dashed: mean of all 17 models. $r$: mean correlation between the model's four task profiles. Rows are
families; post-trained models carry the suffix -Inst, -SFT or -DPO.}
\label{fig:profiles}
\end{figure}

\begin{figure}[H]\centering
\includegraphics[width=\textwidth]{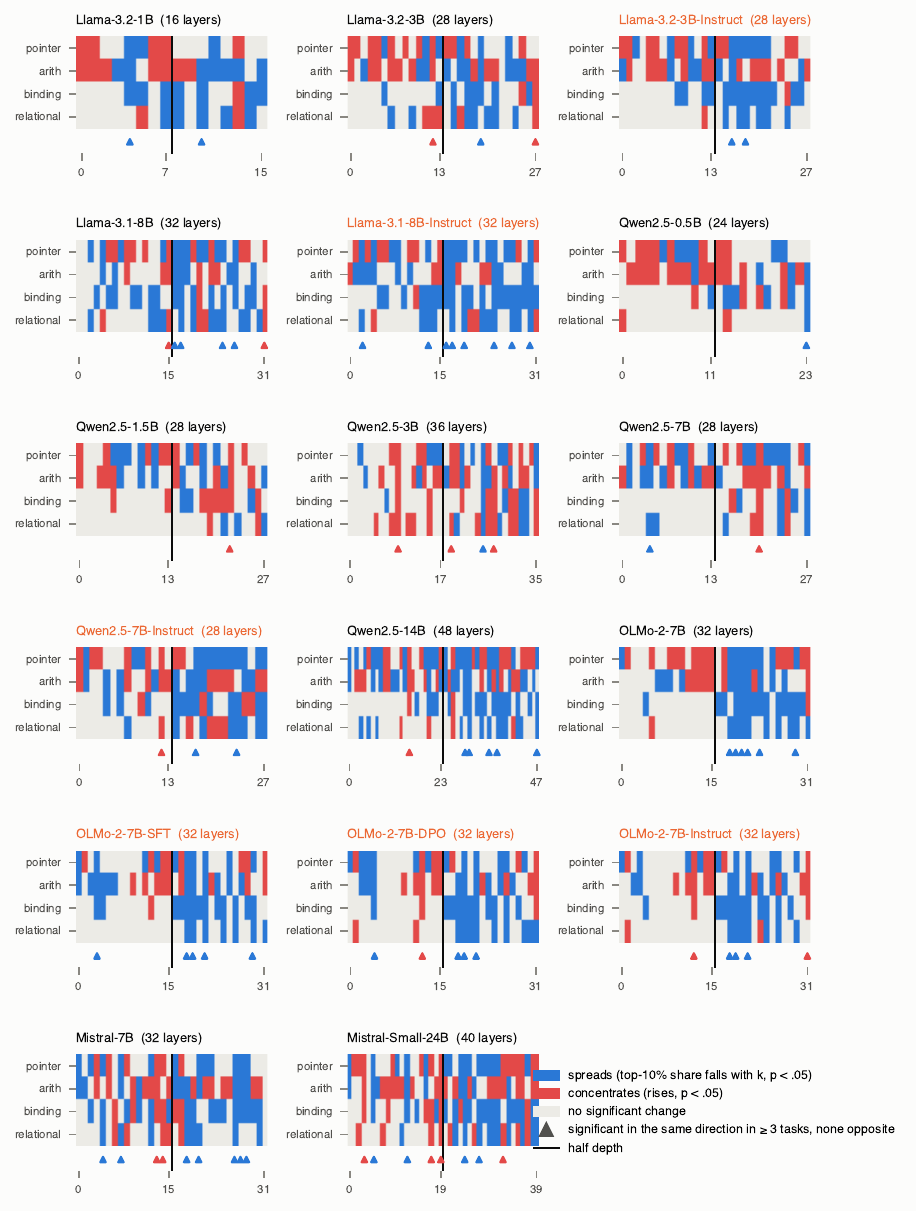}
\caption{\textbf{Layer-by-task maps for every model (top-10\% head share).} Rows: tasks; columns: layers; blue
layers spread ($p<0.05$), red layers concentrate. Triangles: layers significant in the same direction in
at least three tasks and significant in the opposite direction in none. Vertical line: half depth.
Orange titles: post-trained models.}
\label{fig:layermaps}
\end{figure}

\begin{figure}[H]\centering
\includegraphics[width=\textwidth]{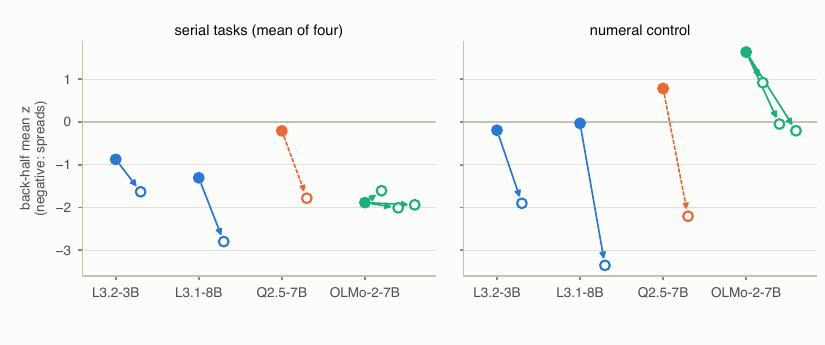}
\caption{\textbf{Back-half response of base and post-trained models.} Base models (filled) and their post-trained versions (open): mean $z$ over back-half layers
in the four serial tasks (left) and in the numeral control (right); negative values mean spreading.
The three open OLMo-2 markers are, from left to right, SFT, DPO and Instruct. Dashed: the Qwen2.5-7B
pair, whose base model was run with eager attention and its Instruct model with SDPA.}
\label{fig:posttrain}
\end{figure}

\end{document}